%% file: main.tex
\documentclass{article}

\makeatletter
\def\input@path{{template/}}
\makeatother

\usepackage{template/fancyhdr}
\usepackage{template/natbib}
\usepackage{preprint,times}

\input{template/math_commands.tex}

\usepackage{amssymb}
\usepackage{amsthm}
\usepackage{algorithm}
\usepackage{algpseudocode}

\theoremstyle{definition}

\usepackage{xcolor}
\usepackage{graphicx}
\graphicspath{{figures/}}
\usepackage{tikz}
\usepackage{pgfplots}
\pgfplotsset{compat=1.18}
\usepgfplotslibrary{groupplots}
\usepackage{booktabs}
\usepackage{longtable}
\usepackage{capt-of}
\usepackage{multirow}
\usepackage{multicol}
\usepackage{array}
\usepackage{adjustbox}
\usetikzlibrary{calc,positioning,arrows.meta,shapes.geometric}

\usepackage{enumitem}
\setlist[itemize]{leftmargin=1.5em}
\setlist[enumerate]{leftmargin=1.5em}
\usepackage{url}
\usepackage{hyperref}
\usepackage[capitalize,noabbrev]{cleveref}

\title{Beyond Best-of-$k$: Prior-Consistent DAPS for Ambiguous Nonlinear Inverse Problems}

\author{
  \begin{minipage}{0.96\textwidth}
  \centering
  \large
  \textbf{Merve Kocabas}\textsuperscript{1,2,3*} \quad
  \textbf{Benjamin A. Burns}\textsuperscript{4*} \quad
  \textbf{Andreas Geiger}\textsuperscript{1,5,6,7} \\[0.25em]
  \textbf{Kent Gauen}\textsuperscript{4} \quad
  \textbf{Sara Fridovich-Keil}\textsuperscript{4} \\[0.65em]
  \normalfont\fontsize{10.5}{12}\selectfont
  \textsuperscript{1}University of T\"ubingen \quad
  \textsuperscript{2}Zuse School ELIZA \\
  \textsuperscript{3}Max Planck Institute for Intelligent Systems \quad
  \textsuperscript{4}Georgia Institute of Technology \\
  \textsuperscript{5}T\"ubingen AI Center \quad
  \textsuperscript{6}ELLIS Institute \quad
  \textsuperscript{7}KE:SAI \qquad
  \textsuperscript{*}Equal contribution \\[0.5em]
  \normalsize
  \texttt{merve.kocabas@student.uni-tuebingen.de} \quad
  \texttt{a.geiger@uni-tuebingen.de} \\
  \texttt{\{bburns46,gauenk,sfk\}@gatech.edu}
  \end{minipage}
  }

\begin{document}

\maketitle

\begin{abstract}
Diffusion models have become strong priors for solving inverse problems. However, nonlinear inverse problems such as phase retrieval remain challenging when multiple measurement-consistent solutions are distinguishable only by the prior; we call such measurements ambiguous. Existing diffusion-based samplers rely on local approximations during reconstruction. For an ambiguous measurement, these approximations can allow measurement guidance to favor a measurement-consistent mode that is close to a prior-supported mode but is itself not supported by the prior. Evaluation often masks this type of failure by reporting the best-of-$k$ runs, where the best is selected by comparison to the ground truth. We introduce Prior-Consistent DAPS (PC-DAPS), a variant of Decoupled Annealing Posterior Sampling (DAPS) designed to address this failure case by maintaining multiple candidate solutions within a single sampling run. As annealing noise decreases, PC-DAPS prunes and replaces weaker candidates according to their prior consistency, without any ground-truth access. PC-DAPS reduces the incidence of prior-inconsistent samples and improves sample quality on nonlinear inverse problems while preserving sample quality and diversity on linear inverse problems.\footnote{Code: \url{https://github.com/voilalab/PC-DAPS};
project website: \href{https://mervekocabas.github.io/PCDAPS/}{PC-DAPS}.}
\end{abstract}

\input{sections/1-introduction}
\input{sections/2-related-work}
\input{sections/3-preliminaries}

\input{sections/4-method}
\input{sections/5-results}
\input{sections/6-discussion}
\input{sections/99-acknowledgements}
\bibliography{refs}
\bibliographystyle{template/preprint}

\clearpage
\appendix
\input{sections/A-appendix}

\end{document}

%% file: template/math_commands.tex
\usepackage{amsmath,amsfonts,bm}

\def\eqref#1{equation~\ref{#1}}

\def\1{\bm{1}}

\DeclareMathAlphabet{\mathsfit}{\encodingdefault}{\sfdefault}{m}{sl}
\SetMathAlphabet{\mathsfit}{bold}{\encodingdefault}{\sfdefault}{bx}{n}

\newcommand{\Var}{\mathrm{Var}}



%% file: sections/1-introduction.tex
\section{Introduction}
\label{sec:introduction}

\begin{figure}[h]
    \centering
    \includegraphics[width=\linewidth]{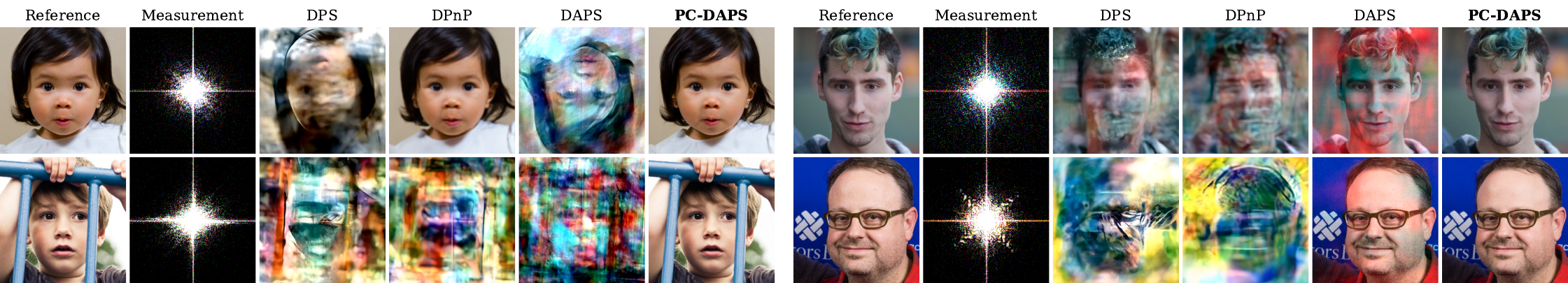}
\caption{\textbf{Measurement consistency is not sufficient for reconstruction.} A single Fourier magnitude measurement is shared by rotations, translations, and color-channel inversions of an image. Existing solvers may return these prior-inconsistent modes, while PC-DAPS selects reconstructions supported by the learned image prior. Each sample was selected randomly, either from four independent runs for other methods or from the four particles generated in a single PC-DAPS run.}
    \label{fig:teaser}
\end{figure}

Inverse problems arise across science and engineering whenever an unknown signal must be recovered from indirect measurements, including image restoration \citep{bertero2021invprobimg}, medical imaging \citep{song2022medical}, and astronomical imaging \citep{event2019first}. Given an observed measurement $y$ generated through a known but potentially noisy measurement process, the goal of the inverse problem is to recover the original signal $x$. For underdetermined problems, in which many different signals $x$ may give rise to the same measurement $y$, Bayesian imaging defines the posterior distribution as $p(x\mid y) \propto p(y\mid x)p(x)$, where the likelihood $p(y\mid x)$ captures measurement consistency of a candidate image $x$ and the prior $p(x)$ captures which candidate reconstructions are plausible natural signals. The goal of posterior sampling is thus to generate candidate reconstructions that are simultaneously consistent with the measurement and the prior.

In Fourier phase retrieval, for example, multiple images can produce identical Fourier magnitude measurements \citep{fienup1982phase}. These measurement-equivalent solutions therefore have the same likelihood, but can have very different posterior probabilities depending on their consistency with the image prior. In particular, we observe that existing phase-retrieval solvers applied to face images sometimes return a rotated or color-inverted face that is consistent with the measurement but implausible under a natural-image prior concentrated on upright faces. The challenge is therefore not simply to find measurement-consistent reconstructions, but to sample the measurement-consistent solutions supported by the prior.

Diffusion models provide strong image priors for posterior sampling \citep{ho2020denoising, Song2021}. At each noise level, a diffusion model implicitly defines a denoiser that maps a noisy sample $x_t$ to an estimate of the corresponding clean image; this diffusion denoiser can be used to incorporate the prior $p(x)$ during posterior sampling. Diffusion-based inverse solvers apply this prior information along a sampling \emph{trajectory}, i.e., the sequence of intermediate states that evolves from a noisy sample toward a reconstruction. Existing methods primarily either tilt the unconditional reverse diffusion process using an approximate likelihood gradient \citep{zhang2025improvingdiffusioninverseproblem, chung2023diffusion, song2023pseudoinverse, boys2024tweedie} or alternate prior and measurement updates across noise levels \citep{wu2024principled, xu2024provably, chen2026daps++, ahmed2026solving}. In these methods, the prior information available at each noise level is represented through a local denoising approximation centered on a single clean-image estimate, rather than explicitly maintaining multiple competing prior modes at intermediate timesteps.

When the measurement is ambiguous, as in the case of phase retrieval and other underdetermined inverse problems, following a sampling trajectory can lead to \emph{premature mode selection}. At high noise levels, multiple prior modes may remain plausible, but the trajectory can only move towards one of them. Once a trajectory selects a particular prior mode, subsequent refinement primarily improves measurement consistency in the local neighborhood of that mode rather than considering more distant alternatives. This becomes problematic when (i) the selected prior mode is not consistent with the measurement, and (ii) its nearest measurement-consistent mode is not supported by the prior. Trajectory-based samplers can prematurely commit to this pair of modes, ultimately resulting in hallucinatory sampled reconstructions that are not supported by the true posterior or even the true prior. Best-of-$k$ evaluation partially masks this issue by generating multiple independent reconstructions and selecting the one closest to the ground truth \citep{chung2023diffusion,
zhang2025improvingdiffusioninverseproblem}. However, this selection uses information unavailable at test time and does not ensure that the produced samples are prior-consistent. Particle-based approaches maintain multiple candidates at intermediate timesteps \citep{cardoso2024monte, dou2024diffusion, wu2023twisteddiffusion}, but have primarily focused on linear inverse problems and particle diversity rather than prior-consistent sampling.

To address this issue, we introduce Prior-Consistent DAPS (PC-DAPS), which moves sample selection inside the sampler and removes the need for ground-truth-based best-of-$k$ evaluation. PC-DAPS maintains a bag of $K$ particles annealed under the same measurement, and models the denoiser posterior $p(x_0 \mid x_t)$ with a $K$-component Gaussian mixture rather than a unimodal estimate. Early in sampling, multiple candidates are kept while the noise level is high and the prior is still weakly informative. As the noise decreases during annealing, particles with low prior consistency are pruned and replaced by surviving candidates that are more prior-consistent. PC-DAPS thus delays mode selection until the prior becomes more informative, performing within a single sampling run (and without ground-truth access) what best-of-$k$ does only after $k$ independent sampling runs.

We evaluate PC-DAPS on FFHQ \citep{karras2019style} and ImageNet \citep{deng2009imagenet} across three nonlinear and seven linear tasks, including the eight tasks reported in DAPS \citep{zhang2025improvingdiffusioninverseproblem} as well as high-pass filtering \citep{thaker2025frequency} and 16$\times$ super-resolution. We report PSNR, SSIM \citep{wang2004image}, and LPIPS \citep{zhang2018unreasonable} under both single-sample and best-of-$k$ evaluation. We additionally compare the rate of prior-inconsistent hallucinations against independent DAPS runs. PC-DAPS reduces the incidence of prior-inconsistent samples and improves performance on ambiguous nonlinear problems while maintaining image quality and diversity on linear inverse problems. Our contributions are as follows:
\begin{itemize}
\item We propose PC-DAPS, which maintains a bag of particles, models the denoiser posterior with a Gaussian mixture, and performs prior-based pruning without access to ground truth.
\item We evaluate PC-DAPS across seven linear and three nonlinear inverse problems on FFHQ and ImageNet, showing fewer prior-inconsistent reconstructions and improved nonlinear reconstruction quality while preserving performance on linear tasks.
\end{itemize}

%% file: sections/2-related-work.tex
\section{Related work}
\label{sec:related-work}

\textbf{Likelihood-guided diffusion sampling.} Given a diffusion prior that has learned the unconditional score $\nabla_{x_t}\log p(x_t)$, we can decompose the conditional score (after observing measurement $y$) as $\nabla_{x_t}\log p(x_t \mid y) = \nabla_{x_t}\log p(x_t) + \nabla_{x_t}\log p(y \mid x_t)$. The likelihood term $\nabla_{x_t}\log p(y \mid x_t)$ is intractable, so a family of methods approximates it to steer the reverse diffusion process. In particular, DPS \citep{chung2023diffusion} evaluates the likelihood at the one-step posterior mean $p(y \mid x_t) \approx p(y \mid \mathbb{E}[x_0 \mid x_t])$ and backpropagates to compute the score. For linear measurement operators, $\Pi$GDM \citep{song2023pseudoinverse} and TMPD \citep{boys2024tweedie} instead propagate Gaussian approximations of $p(x_0 \mid x_t)$, while DDRM \citep{kawar2022denoising} utilizes a diagonalization approach. MCG \citep{chung2022improving} adds a manifold-correction term to keep the guided iterate close to the data manifold. \citet{jin2026hallucination} target hallucination inside this family of approaches by stabilizing the prior update, but remain committed to a single trajectory. 

\textbf{Split and annealed posterior sampling.} In contrast, variable-splitting methods alternate between prior-only and measurement-only steps. Among methods using a diffusion model for the prior step, PnP-DM \citep{wu2024principled} uses a split Gibbs sampler for the likelihood, while DPnP \citep{xu2024provably} uses Metropolis-Hastings and DiffPIR \citep{zhu2023denoising} utilizes a proximal step. Finally, DAPS \citep{zhang2025improvingdiffusioninverseproblem} utilizes a repeated denoising-renoising scheme (\cref{sec:preliminaries-daps}). RePS \citep{ahmed2026solving} adds restarts to that schedule, and DAPS++ \citep{chen2026daps++} reports that the DAPS reconstruction is largely driven by measurement consistency. We therefore design PC-DAPS as a modification of DAPS that retains its variable-splitting structure but introduces prior-based selection to discriminate between measurement-consistent reconstructions.

\textbf{Particle-based posterior sampling.} Another group of approaches maintains many particles throughout sampling, and uses sequential Monte Carlo to propagate, reweight, and resample particles to progressively improve measurement consistency \citep{cardoso2024monte, janati2025mixture}. FPS-SMC \citep{dou2024diffusion} casts the reverse sampling process as a state-space model and applies a particle filter. TDS \citep{wu2023twisteddiffusion} uses twisting functions to obtain asymptotic exactness for general measurement models. While particle-based methods are typically concerned with the fidelity of the ensemble in the limit of many particles, we are concerned instead with the probability that a single returned reconstruction is prior-inconsistent, and use particles for the purpose of exploring the posterior landscape.

\textbf{Diagnosing hallucination and prior misfit.} Hallucination has been studied as a property of reconstruction algorithms in inverse problems since before diffusion priors were adopted \citep{bhadra2021hallucinations}, and refers to any reconstruction that is inconsistent with either the measurement or the prior. Recent work has sought to quantify hallucination specifically for diffusion-based solvers \citep{jin2026hallucination}. At the level of the model rather than the individual sample, DiME \citep{wang2026sample} estimates the evidence of a diffusion prior by integrating over the time-marginals of a posterior sampler, and can be run alongside DAPS to detect prior misfit. Closest to our framing, \citet{burns2026and} adopt a finite-sample perspective on reverse diffusion dynamics and show that widely used guidance approximations misestimate posterior spread at intermediate noise levels. This misestimation produces both prior modes absent from the true posterior and likelihood (measurement-consistent) modes unsupported by the prior, which is the specific failure that our PC-DAPS method aims to address. Benchmarks whose posteriors are analytically tractable provide a complementary check on the uncertainty a solver reports \citep{crafts2025benchmarking, zach2026statistical, zheng2025inversebench}. All of these existing hallucination diagnostics are applied after sampling; PC-DAPS utilizes prior support during sampling to decide which reconstructions survive, thereby reducing the incidence of prior-inconsistent hallucinations.

%% file: sections/3-preliminaries.tex
\section{Preliminaries}
\label{sec:preliminaries}

\subsection{Bayesian inverse problems}
\label{sec:preliminaries-inverse}

We consider Bayesian inverse problems of the form:
\begin{equation}
    y = \mathcal{A}(x) + n, \qquad n \sim \mathcal{N}(0, \beta_y^2 I),
\label{eq:forward-model}
\end{equation}
where $x \in \mathbb{R}^d$ is the unknown signal of interest, $\mathcal{A} : \mathbb{R}^d \to \mathbb{R}^m$ is a known measurement operator that is deterministic but may be linear or nonlinear, $n \in \mathbb{R}^m$ is unknown additive Gaussian noise, and $y \in \mathbb{R}^m$ is the observed measurement. Given a prior distribution $p(x)$, Bayes' rule gives the posterior distribution over likely signals $x$ consistent with a provided measurement $y$:
\begin{equation}
    p(x \mid y) \propto p(y \mid x)\, p(x), \qquad p(y \mid x) = \mathcal{N}\bigl(y; \mathcal{A}(x), \beta_y^2 I\bigr).
\label{eq:posterior}
\end{equation}
We are particularly interested in forward models where the set of measurement-consistent signals
\begin{equation}
    \mathcal{M}(y) = \left\{ x \in \mathbb{R}^d : \|\mathcal{A}(x)-y\| \leq \delta \right\}
\label{eq:consistent-set}
\end{equation}
contains multiple candidate reconstructions which are well-separated in signal space. In this context, two distinct signals in $\mathcal{M}(y)$ can be assigned equal likelihoods $p(y\mid x)$ but have drastically different posterior probabilities $p(x\mid y)$, as one image might be unrealistic (i.e., significantly lower prior probability $p(x)$). A common failure case of existing reconstruction methods is to sample both signals with roughly equal frequency, thereby misrepresenting their posterior probabilities as relatively equal.

For example, in Fourier phase retrieval, two images which are identical up to rotation, translation, and/or color channel sign appear visually quite distinct but produce identical Fourier magnitude measurements \citep{fienup1982phase}. On structured face images, a solver can return a $180^\circ$-rotated or color-inverted (e.g., blue) face that fits the observed Fourier measurement despite being inconsistent with a prior distribution supported only on the set of upright faces.  We refer to such reconstructions as \emph{prior-inconsistent}. Prior-inconsistent reconstructions commonly appear across classical phase-retrieval methods \citep{fienup1982phase}, deep-learning approaches \citep{metzler2018prdeep, icsil2025deep}, and diffusion-based inverse-problem solvers \citep{xu2024provably, zhang2025improvingdiffusioninverseproblem}. Our goal is to reduce the frequency of prior-inconsistent solutions while retaining quality and diversity of measurement- and prior-consistent samples.

\subsection{Decoupled Annealing Posterior Sampling (DAPS)}
\label{sec:preliminaries-daps}

To develop a sampler which produces fewer prior-inconsistent samples, we build on existing approaches by (i) identifying the root causes that produce the inconsistent samples and (ii) making adjustments to reduce the incidence of these samples. Specifically, we build on DAPS \citep{zhang2025improvingdiffusioninverseproblem}, a diffusion-based variable-splitting approach which repeatedly (i) denoises to a clean image via the reverse diffusion process, (ii) improves the clean image's likelihood consistency via MCMC (e.g., Langevin dynamics \citep{roberts1996exponential}), and (iii) renoises the MCMC output via the forward diffusion process. By adding less noise back than was originally present, DAPS aims to iteratively denoise towards a measurement-consistent clean sample.

Mathematically, DAPS utilizes the fact that given an intermediate image $x_t \sim p(x_t \mid y)$, drawing:
\begin{equation}
    x_0 \sim p(x_0 \mid x_t, y), \qquad x_{t-1} \sim \mathcal{N}\!\left(x_0, \sigma_{t-1}^2 I\right),
    \label{eq:daps-marginal}
\end{equation}
produces $x_{t-1} \sim p(x_{t-1} \mid y)$. To sample from the guided denoising distribution $p(x_0 \mid x_t, y)$, observe that $y$ and $x_t$ are independent conditioned on $x_0$; thus, by Bayes' rule we can factorize:

\begin{equation}
    p(x_0 \mid x_t, y) \propto p(y \mid x_0) p(x_0 \mid x_t).
    \label{eq:daps-factorization}
\end{equation}
While $p(y \mid x_0)$ is induced by \cref{eq:forward-model}, the denoising distribution $p(x_0 \mid x_t)$ is unknown and intractable to estimate via sampling. DAPS utilizes a Gaussian approximation for this distribution:
\begin{equation}
    p(x_0 \mid x_t) \approx \mathcal{N}\bigl(x_0; \hat{x}_0(x_t), r_t^2 I\bigr),
    \label{eq:daps-gaussian}
\end{equation}
where $\hat{x}_0(x_t)$ is an estimate of $\mathbb{E}[x_0 \mid x_t]$ and $r_t$ is set heuristically. Observe that the (Stein) score w.r.t. $x_0$ of the \emph{approximated} right-hand side of \cref{eq:daps-factorization} is available in closed form, allowing for sampling via the unadjusted Langevin algorithm with step size $\eta_t$ from initialization $\hat{x}_0(x_t)$:
\begin{equation}
    \hat{x}_0^{(j+1)} \leftarrow \hat{x}_0^{(j)} + \eta_t \bigl( \nabla_{\hat{x}_0} \log p(\hat{x}_0^{(j)} \mid x_t) + \nabla_{\hat{x}_0} \log p(y \mid \hat{x}_0^{(j)}) \bigr) + \sqrt{2\eta_t} \epsilon^{(j)} , \quad \epsilon^{(j)} \sim \mathcal{N}(0, I).
    \label{eq:daps-langevin}
\end{equation}
Once the Langevin algorithm terminates (after a user-prescribed number of iterations $N$), the result is renoised using the EDM parameterization \citep{karras2022elucidating} of the forward diffusion process $x_{t-1} = \hat x_0^{(N)} + \sigma_{t-1} \epsilon$ with $\epsilon \sim \mathcal{N}(0, I)$. Renoising in DAPS is introduced to counteract the locality of Langevin refinement: because consecutive Langevin updates are local, the samples may fail to explore alternative prior modes distant from $\hat x_0(x_t)$. The renoising-denoising update intends to allow particles to make larger moves per iteration and to investigate alternative prior modes.

\begin{figure*}
    \centering
    \begingroup
    \def\figrowdir{figures/gate-toy-exp/}
    \input{figures/gate-toy-exp/gate-toy-exp}
    \endgroup
    \caption{\textbf{Premature mode selection.} We consider a nonlinear inverse problem with quadratic $\mathcal{A}$ and a two-atom prior; details are provided in \cref{app:toy-experiment}. We select $y$ such that one atom is likelihood-consistent while the other is likelihood-inconsistent but neighboring a likelihood-consistent prior-inconsistent mode. The exact posterior, DAPS, and PC-DAPS are shown as distributions across noise levels $\sigma$. The exact posterior assigns positive weight to both atoms, with majority weight on the likelihood-consistent atom. However, DAPS assigns majority mass to the prior-inconsistent likelihood mode, as highlighted in the last panel, which is not present in the true posterior. PC-DAPS's selection algorithm prunes this mode, sampling only from the posterior-consistent atom. While neither DAPS nor PC-DAPS samples from the likelihood-inconsistent atom with adequate probability, PC-DAPS eliminates the hallucinated prior-inconsistent mode produced by DAPS.}
    \label{fig:toy-example}
\end{figure*}
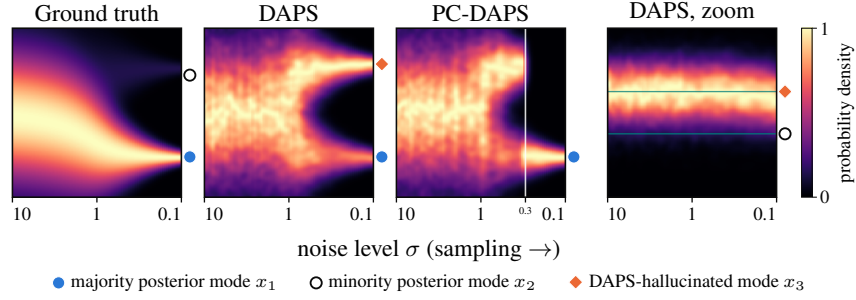

However, the DAPS update rule still produces the prior-inconsistent samples described in \cref{sec:preliminaries-inverse}, as shown in \cref{fig:teaser}. Observe that the DAPS procedure for approximately sampling from $p(x_0 \mid x_t, y)$ in effect produces a prior-consistent sample near $\mathbb{E}[x_0 \mid x_t]$ and samples from likelihood-consistent regions of its neighborhood. In contexts where $x_t$ selects a prior-consistent sample that is likelihood-inconsistent, yet there exists a nearby region of signal space consistent with $y$, \cref{eq:daps-langevin} may sample from that region \emph{even if it is prior-inconsistent}, as the Gaussian approximation to $p(x_0 \mid x_t)$ merely discourages the samples from deviating far in signal space from $\hat x_0(x_t)$. We observe that $x_0$ renoises to an $x_t$ that subsequently denoises to the same region with high probability, causing the particle to become trapped in a prior-inconsistent but measurement-consistent mode. We refer to this phenomenon as \emph{premature mode selection} because the sample $x_t$ prematurely commits to only sampling $x_0$ from a particular prior mode, without being able to adequately decommit even if that prior mode is not present in the posterior.

\Cref{fig:toy-example} illustrates this phenomenon. Here, the prior places mass on two candidate solutions, $x_1$ and $x_2$, while the measurement $y$ is explained by $x_1$ and a third point $x_3$ unsupported by the prior. In analogy to Fourier phase retrieval, one can think of $x_1$ as the desired reconstruction, e.g., an upright, non-blue face that is both consistent with the measurement and supported by the image prior, $x_2$ as a realistic face inconsistent with the particular measurement, and $x_3$ as a structurally implausible transformation of the desired image $x_1$, such as a rotated or blue face. The exact posterior should favor the $x_1$ mode while assigning zero mass to the prior-inconsistent $x_3$. However, DAPS samples from $x_3$'s neighborhood by (i) sampling $x_2$ from the prior and (ii) sampling $x_3$ from $x_2$'s local likelihood. Once a particle enters this locally attractive likelihood-consistent region, subsequent renoising and Langevin updates become trapped, resulting in premature mode selection. 

This toy example motivates the design of PC-DAPS, which maintains multiple candidate measurement-consistent modes to identify and correct trapped prior-inconsistent samples. As shown in \cref{fig:toy-example}, PC-DAPS places its mass only on the prior-supported and measurement-consistent mode at $x_1$ and avoids sampling from the prior-inconsistent, measurement-consistent region near $x_3$. We describe how PC-DAPS identifies and removes such prior-inconsistent particles in the next section.

%% file: figures/gate-toy-exp/gate-toy-exp.tex
\providecommand{\figrowdir}{}
\definecolor{rowBlue}{HTML}{2A78D6}
\definecolor{rowOrange}{HTML}{EB6834}
\definecolor{rowInk}{HTML}{0B0B0B}
\definecolor{rowGray}{HTML}{52514E}
\definecolor{rowTeal}{HTML}{007C7A}
\begin{tikzpicture}[
    fits/.style={circle, fill=rowBlue, draw=white, line width=0.3pt, inner sep=0pt, minimum size=4.6pt},
    nofit/.style={circle, fill=white, draw=rowInk, line width=0.7pt, inner sep=0pt, minimum size=4.2pt},
    offsupport/.style={diamond, fill=rowOrange, draw=white, line width=0.3pt, inner sep=0pt, minimum size=5.4pt},
    ptitle/.style={anchor=south, font=\footnotesize, inner sep=0pt},
    psub/.style={anchor=south, font=\scriptsize, text=rowGray, inner sep=0pt}]
\pgfplotsset{
  rowframe/.style={scale only axis, height=0.88in, anchor=south west, axis on top, clip=false,
    enlargelimits=false, axis line style={line width=0.6pt, rowInk}, ytick=\empty},
  rowheat/.style={rowframe, width=0.88in, xmin=0, xmax=1, xtick={0,0.5,1}, xtick pos=lower,
    xticklabels={\makebox[0pt][l]{10}, 1, \makebox[0pt][r]{0.1}},
    xticklabel style={font=\scriptsize, inner sep=1.5pt}, tick align=outside, major tick length=2pt,
    every tick/.style={rowInk, line width=0.5pt}}}

\begin{axis}[rowheat, at={(0.98in,0in)}, ymin=-1.9, ymax=1.9]
  \addplot graphics[xmin=0, xmax=1, ymin=-1.9, ymax=1.9] {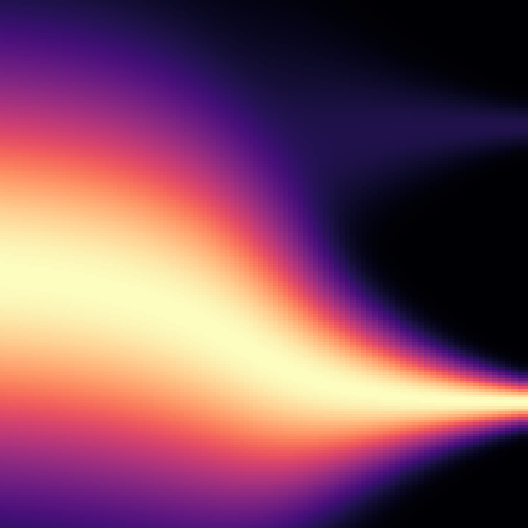};
  \node[fits] at (axis cs:1.05,-0.9944) {};
  \node[nofit, yshift=-2.5pt] at (axis cs:1.05,0.9944) {};
\end{axis}
\node[ptitle] at (1.42in,0.915in) {Ground truth};

\begin{axis}[rowheat, at={(1.98in,0in)}, ymin=-1.9, ymax=1.9]
  \addplot graphics[xmin=0, xmax=1, ymin=-1.9, ymax=1.9] {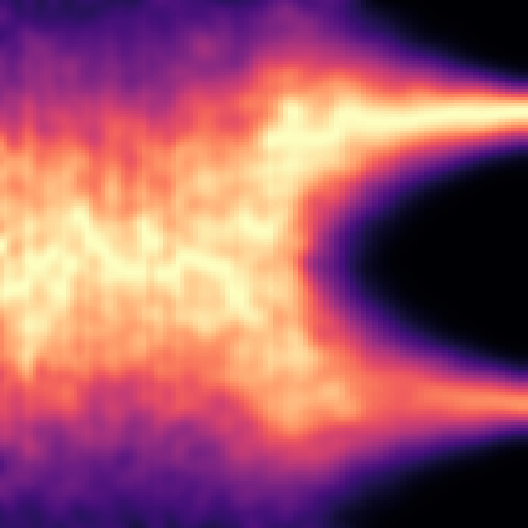};
  \node[fits] at (axis cs:1.05,-0.9944) {};
  \node[offsupport] at (axis cs:1.05,1.0939) {};\end{axis}
\node[ptitle] at (2.42in,0.915in) {DAPS};

\begin{axis}[rowheat, at={(2.98in,0in)}, ymin=-1.9, ymax=1.9,
    extra x ticks={0.7636}, extra x tick labels={\scalebox{0.55}{0.3}}]
  \addplot graphics[xmin=0, xmax=1, ymin=-1.9, ymax=1.9] {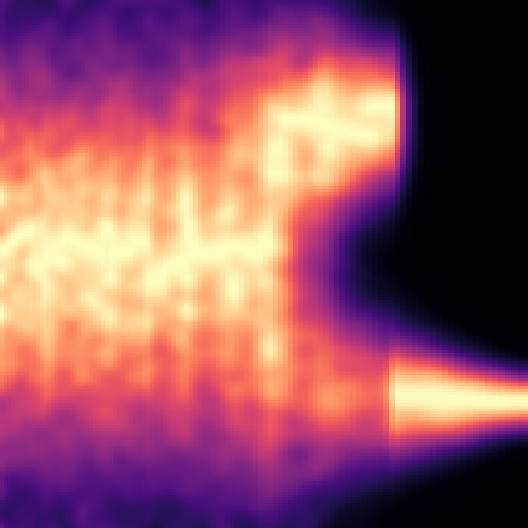};
  \draw[white, line width=0.6pt, opacity=0.85] (axis cs:0.7636,-1.9) -- (axis cs:0.7636,1.9);
  \node[fits] at (axis cs:1.05,-0.9944) {};
\end{axis}
\node[ptitle] at (3.42in,0.915in) {PC-DAPS};

\begin{axis}[rowheat, at={(4.08in,0in)}, ymin=0.8500, ymax=1.2500]
  \addplot graphics[xmin=0, xmax=1, ymin=0.8500, ymax=1.2500] {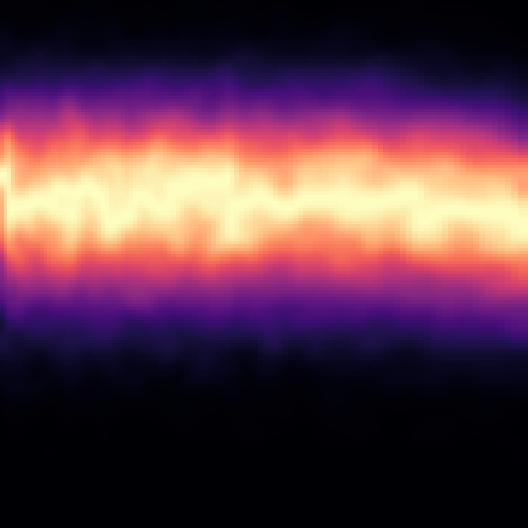};
  \draw[rowTeal, line width=0.5pt, opacity=0.9] (axis cs:0,1) -- (axis cs:1,1);
  \draw[rowTeal, line width=0.5pt, opacity=0.75] (axis cs:0,1.1) -- (axis cs:1,1.1);
  \node[nofit] at (axis cs:1.05,1) {};
  \node[offsupport] at (axis cs:1.05,1.1) {};
\end{axis}
\node[ptitle] at (4.52in,0.915in) {DAPS, zoom};

\node[anchor=south west, inner sep=0pt] at (5.09in,0in)
  {\includegraphics[width=0.055in, height=0.88in]{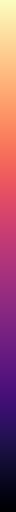}};
\draw[rowInk, line width=0.4pt] (5.09in,0in) rectangle (5.145in,0.88in);
\draw[rowInk, line width=0.4pt] (5.145in,0in) -- ++(0.025in,0in) node[right, font=\scriptsize, inner sep=1pt] {0};
\draw[rowInk, line width=0.4pt] (5.145in,0.88in) -- ++(0.025in,0in) node[right, font=\scriptsize, inner sep=1pt] {1};
\node[rotate=90, font=\scriptsize] at (5.32in,0.44in) {probability density};

\node[anchor=north, font=\footnotesize] at (3.15in,-0.17in) {noise level $\sigma$ (sampling $\rightarrow$)};
\node[anchor=north, font=\scriptsize, inner sep=0pt] at (3.15in,-0.40in) {%
  \tikz[baseline=(blueLabel.base)]{
    \node[fits] (blueMarker) {};
    \node[anchor=west, align=center, inner sep=0pt] (blueLabel)
      at ([xshift=2pt]blueMarker.east) {majority posterior mode $x_1$};
  }\hspace{12pt}%
  \tikz[baseline=(whiteLabel.base)]{
    \node[nofit] (whiteMarker) {};
    \node[anchor=west, align=center, inner sep=0pt] (whiteLabel)
      at ([xshift=2pt]whiteMarker.east) {minority posterior mode $x_2$};
  }\hspace{12pt}%
  \tikz[baseline=(orangeLabel.base)]{
    \node[offsupport] (orangeMarker) {};
    \node[anchor=west, align=center, inner sep=0pt] (orangeLabel)
      at ([xshift=2pt]orangeMarker.east) {DAPS-hallucinated mode $x_3$};
  }};
  
\end{tikzpicture}

%% file: sections/4-method.tex
\section{Method}
\label{sec:method}

To address the presence of prior-inconsistent samples, a first approach might be to produce multiple samples via DAPS and discard any which are prior-inconsistent. There are two main shortcomings of this approach: (i) the effective sample size of this sampling method is suboptimal, as substantial computational resources are used to produce samples that are then thrown out, and (ii) it is unclear how to select which samples to discard without having access to the ground truth signal or posterior. 

To address these issues, we introduce PC-DAPS, a procedure for correcting prematurely committed particles \emph{during the sampling process} without requiring access to the ground truth image or to any pre-trained prior-inconsistency classifier. PC-DAPS maintains a bag of $K$ particles, each of which is denoised using the same annealing schedule and pretrained denoiser, but uses different initializations and random noise in its Langevin and renoising updates. If no additional modifications are made to DAPS, the $K$ particles will denoise independently, and can be partitioned into (nonempty) sets of ``bad'' prematurely committed particles and ``good'' particles that are posterior-consistent. The core idea of PC-DAPS is to improve DAPS's existing renoising strategy by identifying prematurely committed particles \emph{during sampling} and moving them to more promising modes discovered by other particles in the bag. Additionally, PC-DAPS utilizes the particles to refine additional design decisions made by DAPS, such as by replacing the unimodal Gaussian approximation to the denoising distribution with an adaptive Gaussian mixture model (GMM) approximation. A toy example illustrating the benefit of this Gaussian mixture approximation is shown in \cref{app:mixture}.

What remains is to describe how PC-DAPS (i) identifies when particles commit prematurely, (ii) identifies which particles are problematic, (iii) corrects the identified particles, and (iv) constructs the multimodal denoising distribution approximation from the bag.
PC-DAPS makes two core changes to DAPS, utilizing $K$ particles annealed in parallel: a pruning technique to correct particles trapped in prior-inconsistent likelihood modes in tasks with likelihood ambiguity (\cref{sec:method-pruning}), and a multimodal denoising distribution approximation estimated from the diversity of the particles (\cref{sec:method-mixture}). Additional implementation details and ablations are provided in \cref{app:implementation}, such as omitting the final Langevin refinement and adaptive ODE stepping from high to low noise to improve reconstruction quality while reducing computational cost (\cref{app:time-cost}). The full algorithm (\cref{alg:pcdaps}) can be found in \cref{app:algorithm}.

\subsection{Pruning scheme}
\label{sec:method-pruning}

PC-DAPS begins by initializing a bag of $K$ particles $\mathcal{B} = \{x^{(k)}_{t_{\max}}\}_{k = 1}^K$ from the initial distribution of the reverse diffusion process, and denoises the particles as prescribed by DAPS. On tasks where $\mathcal{M}(y)$ (\cref{eq:consistent-set}) contains multiple clearly separated candidates (e.g., for a nonlinear $\mathcal{A}$), the particles will begin to denoise to genuinely different candidate reconstructions. At the annealing time of the DAPS denoising process where the prior first becomes discriminative, PC-DAPS invokes a one-shot consensus check to identify prior-inconsistent particles.

To determine when the check should occur, we test the diversity of the bag at each iteration. Define:
\begin{equation}
    \bar{r} = \frac{1}{K}\!\sum_{k=1}^{K} \left\|\mathcal{A}(x^{(k)}) - y\right\|^2, \qquad r^{\star} = \big\|\mathcal{A}(\bar{x}) - y\big\|^2, \qquad \bar{x} = \frac{1}{K}\sum_{k=1}^K x^{(k)}.
    \label{eq:pcdaps-jensen-quantities}
\end{equation}
Observe that, for linear $\mathcal{A}$, Jensen's inequality forces $\bar{r} \geq r^{\star}$ by convexity of the squared norm. Using this convexity intuition, for nonlinear $\mathcal{A}$ the condition that $\bar r < r^\star$ indicates that the bag occupies multiple wells of the likelihood potential, indicating that mode-level pruning is meaningful.

To identify which particles require correcting, we partition the bag into clusters $\mathcal{C}_m$ and evaluate each by its average prior consistency (obtained via Tweedie's formula) over several noise scales $\mathcal{S}$:
\begin{equation}
    s(\mathcal{C}_m) = -\frac 1{|\mathcal{C}_m||\mathcal{S}|} \sum_{x \in \mathcal{C}_m} \sum_{\sigma \in \mathcal{S}} \bigl\|x - D_\theta(x + \sigma\epsilon;\, \sigma)\bigr\|^2, \qquad \epsilon \sim \mathcal{N}(0, I).
    \label{eq:pcdaps-prior-score}
\end{equation}
The premise of this criterion is that a sample lying near the (prior) data manifold denoises with small (Stein) score at every noise scale. If a cluster's prior-consistency score $s(\mathcal{C}_m)$ exceeds a threshold, its particles are unaltered. Otherwise, its particles are replaced randomly by copies of particles from clusters scoring above the threshold. Implementation choices including the clustering method, choice of noise scales $\mathcal{S}$, and determination of threshold are discussed in \cref{app:implementation}.

\subsection{Gaussian mixture approximation of the denoiser posterior}
\label{sec:method-mixture}

An additional benefit of maintaining a bag of particles is that we can improve upon DAPS's unimodal approximation in \cref{eq:daps-gaussian} by selecting a multimodal distribution informed by the bag's diversity. We propose replacing the DAPS Gaussian approximation with a Gaussian mixture. Denote the clean-signal estimate of the $k$-th particle by
$\tilde{x}_0^{(k)} = D_\theta\bigl(x_t^{(k)};\, \sigma_t\bigr)$.
Before the pruning described in \cref{sec:method-pruning} activates, we use a uniformly-weighted mixture of Gaussians centered at the clean estimates:
\begin{equation}
    q(x_0 \mid x_{t}) = \frac1K\sum_{k=1}^{K} \mathcal{N}\!\left(x_0; \tilde x_0^{(k)}, \mathrm{diag}\left(\mathrm{Var}_{\mathcal{B}}\big(\tilde{x}_0^{(k)}\big)\right)\right).
\end{equation}
After pruning, we maintain a mode for each unpruned \emph{cluster}, where the corresponding component parameters are given by the cluster's empirical weight, mean, and diagonal covariance:
\begin{gather}
    q(x_0 \mid x_{t}) = \sum_{m=1}^{M}\pi_m\, \mathcal{N} \big(x_0;\mu_m,\Sigma_m\big),
    \label{eq:pcdaps-mixture} \\
    \pi_m = \frac{|\mathcal{C}_m|}{K},
    \qquad
    \mu_m = \frac{1}{|\mathcal{C}_m|}\sum_{x^{(k)} \in \mathcal{C}_m} \tilde{x}_0^{(k)},
    \qquad
    \Sigma_m = \mathrm{diag}\!\left(\mathrm{Var}_{\mathcal{C}_m}\!\big(\tilde{x}_0^{(k)}\big)\right).
    \label{eq:pcdaps-mixture-params}
\end{gather}
When a cluster contains only a single particle, its component uses the full bag's diagonal covariance. 

Unlike DAPS, which centers the denoising approximation at the point estimate $\hat x_0(x_t)$, PC-DAPS shares the bag-mean center across all particles in a mode, giving a variance-reduced prior estimate that we find improves reconstruction quality on linear reconstruction tasks. Furthermore, when the full bag lies in a single cluster, the mixture reduces to a single Gaussian centered on the bag's average Tweedie estimate with data-dependent covariance. This contrasts with the heuristic, geometry-independent covariance utilized by DAPS.

Observe that the score of the GMM denoising approximation $q(x_0 \mid x_t)$ is accessible in closed form:
\begin{equation}
    \nabla_{x_0}\log q(x_0 \mid x_t) = \sum_{m=1}^{M} q_m(x_0) \Sigma_m^{-1} \big(\mu_m - x_0\big),
    \label{eq:pcdaps-mixture-score}
\end{equation}
with weights $q_m(x_0) = \pi_m \mathcal{N}(x_0; \mu_m, \Sigma_m) / \sum_{m'} \pi_{m'} \mathcal{N}(x_0; \mu_{m'}, \Sigma_{m'})$, and thus can be utilized in the DAPS Langevin update introduced in \cref{eq:daps-langevin}. Observe that when the weights $q_m$ are approximately one-hot, this update approximately reduces to a data-dependent Gaussian. Recalling that the DAPS Gaussian approximation disincentivizes the Langevin dynamics from deviating far from the point estimate $\hat x_0(x_t)$, the PC-DAPS Gaussian approximation discourages the Langevin dynamics from deviating far from the \emph{prior mode} identified by a collective cluster of the bag, as demonstrated in \cref{fig:gmm-toy} in \cref{app:mixture}.

%% file: sections/5-results.tex
\section{Results}
\label{sec:results}

\paragraph{Main results.}
\label{sec:results-main}

\input{tables/main-results}

\begin{figure*}[t]
    \centering
    \begin{minipage}[t]{0.48\textwidth}
        \vspace{0pt}
        \centering
        \input{tables/phase-retrieval-outcomes}
    \end{minipage}
    \hfill
     \begin{minipage}[t]{0.5\textwidth}
        \vspace{0pt}
        \centering
        \includegraphics[width=\linewidth]{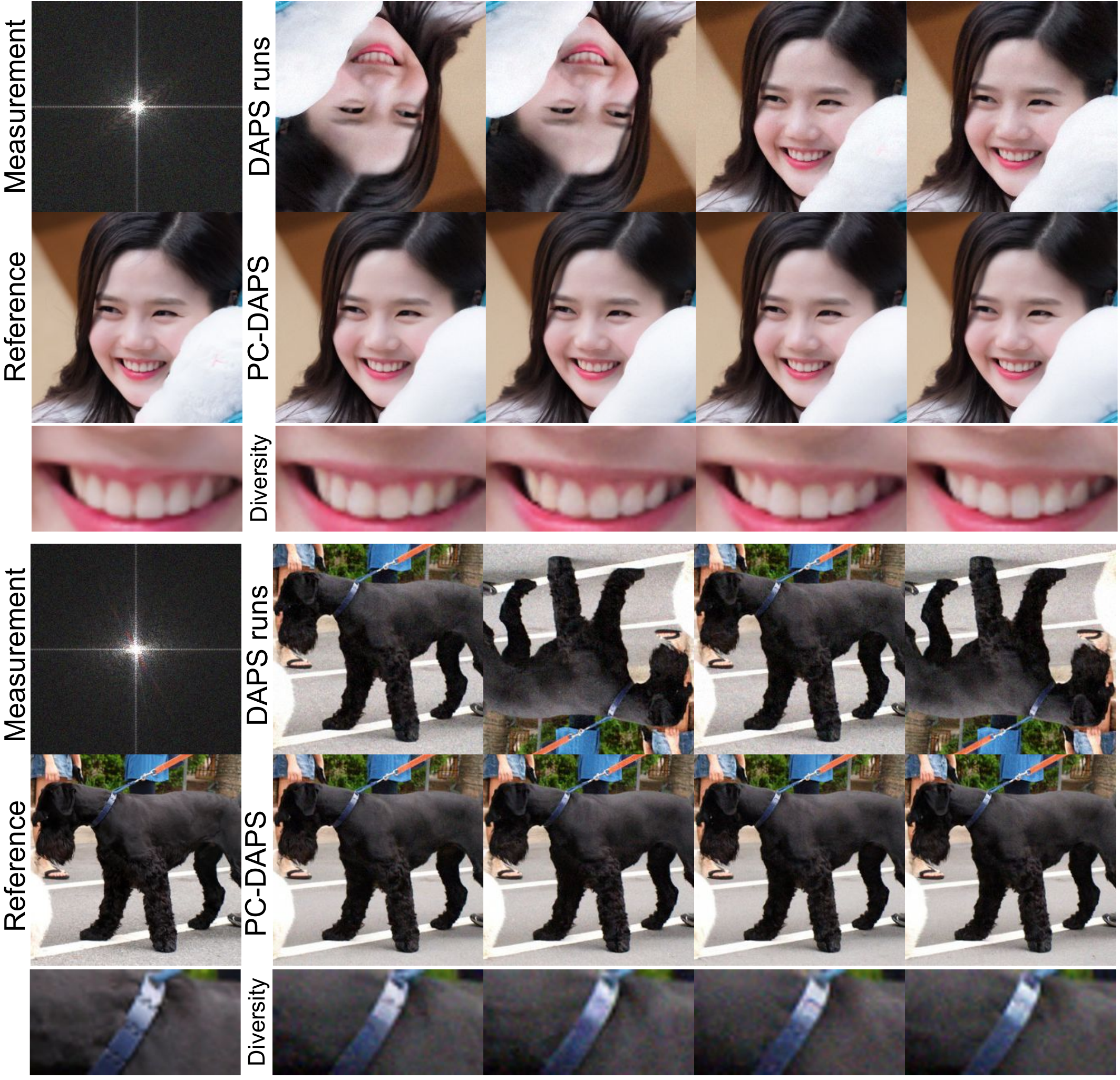}
        \caption{\textbf{PC-DAPS resolves the upside-down orientation inconsistencies seen in standard DAPS best-of-$k$ runs.} The diversity rows show that, although PC-DAPS is stable, it preserves subtle variation among the four particles generated within a single run in comparison to the four independent DAPS runs.}
        \label{fig:phase-retrieval}
     \end{minipage}
\end{figure*}

\Cref{tab:main-results} shows that while PC-DAPS is not uniformly superior to all baselines, it performs competitively across all experiments and often achieves the highest sample quality, with the most substantial improvements in phase retrieval. In particular, the phase-retrieval results for PC-DAPS are competitive across metrics despite its handicap in evaluation protocol; for PC-DAPS we average metrics across the four samples it generates in a single run, whereas all baselines are evaluated on only the best sample (chosen using ground-truth knowledge) generated across four independent runs.
Moreover, despite it's prune-and-replace strategy, we observe that the samples generated within a single PC-DAPS bag typically still contain a similar level of diversity as samples generated by independent DAPS runs, as shown in \cref{fig:phase-retrieval} and \cref{app:more-samples}.

\paragraph{Prior-inconsistent hallucinations.}
\label{sec:results-hallucinations}

\Cref{tab:main-results} reports standard metrics to facilitate baseline comparisons. However, the evaluation procedure masks the failure mode PC-DAPS aims to address (measurement-consistent reconstructions that the prior does not support), because the best-of-$k$ evaluation protocol used by baseline methods discards $k-1$ of the reconstructed images. On phase retrieval, we instead classify every sample as \emph{faithful}, \emph{upside-down / color-inverted} (a measurement-consistent likelihood mode the natural-image prior would down-weight), or \emph{failed}. Labels are assigned automatically using per-channel PSNR to the ground truth and its rotated version, maximizing over $\pm1$-pixel circular shifts to account for translation ambiguity. A channel is assigned to an orientation only when it is more than $1$ dB closer to that orientation; samples with all channels matching the ground truth are \emph{faithful}, those matching the rotation or disagreeing across channels are \emph{upside-down / color-inverted}, and the remainder are \emph{failed}. \Cref{tab:phase-retrieval-outcomes} reports the resulting frequencies, with examples in \cref{fig:phase-retrieval}.

On FFHQ, PC-DAPS lowers the rate of upside-down or color-inverted reconstructions from $14.5\%$ to $1.5\%$ and raises the rate of faithful reconstructions from $79.5\%$ to $97.5\%$. This addresses the main failure mode PC-DAPS is meant to fix. On ImageNet, the improvements are smaller: the proportion of faithful reconstructions increases from $38.5\%$ to $45.0\%$, while the proportion of unrecoverable runs/bags (runs/bags where there exists no faithful sample) drops from $25\%$ to $15\%$, and the rate of upside-down or color-inverted reconstructions decreases from $42.0\%$ to $30.5\%$. PC-DAPS relies on prior consistency to disambiguate the color and orientation ambiguities in phase-retrieval measurements, but the multi-class ImageNet prior provides weaker disambiguating signals at varying noise scales when compared to the FFHQ prior. When the local geometry of the prior support is more complex, data-manifold-based prior-consistency pruning appears to be less effective.

%% file: tables/main-results.tex
\begin{table*}[t]
\centering
\caption{\textbf{Quantitative evaluation on FFHQ and ImageNet.} Comparison of methods on FFHQ (left) and ImageNet (right) across seven linear and three nonlinear inverse problems (defined in \cref{app:inverse-setup}) using PSNR, SSIM, and LPIPS, evaluated over 100 test images. For the nonlinear problems, mean $\pm$ standard deviation over 100 images is reported to highlight variation across samples. PC-DAPS produces 4 samples for each input, so the metrics reported for PC-DAPS are averaged over these 4 samples. For phase retrieval, for all methods other than PC-DAPS, we report oracle results from best-of-4 runs. We indicate the best and second-best results for each experiment with \textbf{bold} and \underline{underline} annotations. We use a noisy measurement with noise level $\beta_y = 0.05$ in all tasks. Baseline implementation details are provided in \cref{app:baseline-details}.}
\label{tab:main-results}
\resizebox{\textwidth}{!}{%
\begin{tabular}{l!{\vrule}l!{\vrule}ccc!{\vrule}ccc}
\toprule
\multirow{2}{*}{\textbf{Task}} & \multirow{2}{*}{\textbf{Method}} & \multicolumn{3}{c}{\textbf{FFHQ}} & \multicolumn{3}{c}{\textbf{ImageNet}} \\  

 &  &  \textbf{PSNR ($\uparrow$)} & \textbf{SSIM ($\uparrow$)} & \textbf{LPIPS ($\downarrow$)} & \textbf{PSNR ($\uparrow$)} & \textbf{SSIM ($\uparrow$)} & \textbf{LPIPS ($\downarrow$)} \\ \midrule

\multirow{5}{*}{Super-resolution $4\times$} 
 & PC-DAPS & $\textbf{30.25}$ & $\textbf{0.841}$ & $\textbf{0.166}$ & $\textbf{26.44}$ & $\textbf{0.714}$ & $\textbf{0.245}$ \\

 & DAPS & $\underline{29.36}$ & $0.782$ & $\underline{0.173}$ & $\underline{25.51}$ & $0.639$ & $\underline{0.281}$ \\
 & DDNM & $28.01$ & $0.797$ & $0.196$ & $23.92$ & $0.606$ & $0.472$ \\

 & DPS & $25.86$ & $0.753$ & $0.269$ & $21.13$ & $0.489$ & $0.361$ \\
 & FPS-SMC & $28.45$ & $\underline{0.812}$ & $0.206$ & $24.79$ & $\underline{0.705}$ & $0.311$ \\

\midrule

\multirow{4}{*}{Super-resolution $16\times$} 
 & PC-DAPS & $\textbf{22.21}$ & $\textbf{0.623}$ & $\textbf{0.368}$ & $\textbf{20.90}$ & $\textbf{0.454}$ & $\textbf{0.535}$ \\

 & DAPS & $\underline{22.14}$ & $0.597$ & $\underline{0.373}$ & $\underline{20.35}$ & $\underline{0.450}$ & $\underline{0.545}$ \\
 & DPS & $21.09$ & $0.586$ & $0.570$ & $18.41$ & $0.384$ & $0.592$ \\
 & FPS-SMC & $22.09$ & $\underline{0.604}$ & $0.496$ & $18.88$ & $0.423$ & $0.588$ \\
\midrule

\multirow{4}{*}{High-pass filter} 
 & PC-DAPS & $\textbf{18.84}$ & $\textbf{0.815}$ & $\textbf{0.155}$ & $\textbf{17.33}$ & $\textbf{0.778}$ & $\textbf{0.204}$ \\

 & DAPS & $16.85$ & $0.702$ & $0.204$ & $\underline{15.51}$ & $\underline{0.732}$ & $\underline{0.217}$ \\
 & DPS & $4.92$ & $0.437$ & $0.687$ & $12.84$ & $0.606$ & $0.473$ \\
 & FGPS & $\underline{18.04}$ & $\underline{0.761}$ & $\underline{0.184}$ & $14.77$ & $0.708$ & $0.229$ \\
\midrule

\multirow{5}{*}{Inpaint (box)} 
 & PC-DAPS & $\underline{24.73}$ & $\underline{0.825}$ & $\underline{0.152}$ & $21.56$ & $\underline{0.738}$ & $\underline{0.221}$ \\

 & DAPS & $24.69$ & $0.746$ & $0.166$ & $21.12$ & $0.714$ & $0.229$ \\
 & DDNM & $24.44$ & $\textbf{0.839}$ & $0.233$ & $\underline{21.67}$ & $\textbf{0.746}$ & $0.321$ \\
 & DPS & $22.51$ & $0.792$ & $0.209$ & $18.94$ & $0.722$ & $0.257$ \\
 & FPS-SMC & $\textbf{24.86}$ & $0.823$ & $\textbf{0.146}$ & $\textbf{22.16}$ & $0.726$ & $\textbf{0.208}$ \\

\midrule

\multirow{5}{*}{Inpaint (random)} 
 & PC-DAPS & $\textbf{31.83}$ & $\textbf{0.875}$ & $\textbf{0.121}$ & $\underline{28.18}$ & $\underline{0.781}$ & $\textbf{0.159}$ \\

 & DAPS & $\underline{30.73}$ & $0.801$ & $0.129$ & $27.32$ & $0.725$ & $\underline{0.189}$ \\
 & DDNM & $29.87$ & $0.816$ & $\underline{0.123}$ & $\textbf{31.14}$ & $\textbf{0.839}$ & $0.193$ \\
 & DPS & $25.46$ & $0.823$ & $0.203$ & $23.52$ & $0.745$ & $0.297$ \\
 & FPS-SMC & $28.17$ & $\underline{0.824}$ & $0.258$ & $24.55$ & $0.699$ & $0.318$ \\

\midrule

\multirow{5}{*}{Gaussian deblurring} 
 & PC-DAPS & $\textbf{30.35}$ & $\textbf{0.839}$ & $\textbf{0.164}$ & $\underline{26.37}$ & $\textbf{0.703}$ & $\textbf{0.250}$ \\

 & DAPS & $\underline{29.60}$ & $0.787$ & $\underline{0.179}$ & $25.94$ & $0.659$ & $\underline{0.267}$ \\
 & DDNM & $28.18$ & $\underline{0.805}$ & $0.214$ & $\textbf{28.11}$ & $\underline{0.701}$ & $0.281$ \\
 & DPS & $25.87$ & $0.764$ & $0.219$ & $20.31$ & $0.598$ & $0.397$ \\
 & FPS-SMC & $26.47$ & $0.771$ & $0.251$ & $23.86$ & $0.603$ & $0.390$ \\

\midrule

\multirow{4}{*}{Motion deblurring} 
 & PC-DAPS & $\textbf{32.13}$ & $\textbf{0.872}$ & $\textbf{0.119}$ & $\textbf{29.06}$ & $\textbf{0.800}$ & $\textbf{0.162}$ \\

 & DAPS & $\underline{31.65}$ & $\underline{0.837}$ & $\underline{0.137}$ & $\underline{28.66}$ & $\underline{0.770}$ & $\underline{0.175}$ \\
 & DPS & $24.52$ & $0.801$ & $0.246$ & $18.96$ & $0.629$ & $0.423$ \\
 & FPS-SMC & $27.31$ & $0.825$ & $0.229$ & $24.60$ & $0.645$ & $0.329$ \\

\midrule

\multirow{7}{*}{Phase retrieval} 
 & PC-DAPS & $\textbf{30.09}_{\pm 1.77}$ & $\textbf{0.834}_{\pm 0.089}$ & $\textbf{0.152}_{\pm 0.088}$ & $\textbf{21.75}_{\pm 6.10}$ & $\underline{0.469}_{\pm 0.132}$ & $\textbf{0.373}_{\pm 0.128}$ \\

 & DAPS & $\underline{29.66}_{\pm 4.17}$ & $\underline{0.770}_{\pm 0.100}$ & $\underline{0.177}_{\pm 0.090}$ & $\underline{20.98}_{\pm 6.72}$ & $0.461_{\pm 0.226}$ & $\underline{0.387}_{\pm 0.163}$ \\
 & DPS & $17.64_{\pm 2.97}$ & $0.441_{\pm 0.129}$ & $0.410_{\pm 0.090}$ & $16.81_{\pm 3.61}$ & $0.427_{\pm 0.143}$ & $0.447_{\pm 0.099}$ \\
 & PnP-DM & $21.70_{\pm 5.50}$ & $0.542_{\pm 0.122}$ & $0.398_{\pm 0.101}$ & $17.68_{\pm 6.70}$ & $0.458_{\pm 0.191}$ & $0.499_{\pm 0.101}$ \\
 & DPnP & $20.39_{\pm 6.15}$ & $0.592_{\pm 0.139}$ & $0.392_{\pm 0.173}$ & $16.74_{\pm 3.21}$ & $\textbf{0.487}_{\pm 0.196}$ & $0.535_{\pm 0.107}$ \\
 & RED-Diff & $15.72_{\pm 4.41}$ & $0.401_{\pm 0.191}$ & $0.588_{\pm 0.089}$ & $15.11_{\pm 3.68}$ & $0.389_{\pm 0.058}$ & $0.529_{\pm 0.127}$ \\

\cmidrule{2-8}
 & HIO & $13.53_{\pm 2.50}$ & $0.359_{\pm 0.093}$ & $0.726_{\pm 0.068}$ & - & - & - \\
\midrule

\multirow{4}{*}{Nonlinear deblur} 
 & PC-DAPS & $\underline{29.04}_{\pm 0.82}$ & $\textbf{0.804}_{\pm 0.017}$ & $\underline{0.180}_{\pm 0.023}$ & $\underline{27.63}_{\pm 2.58}$ & $\underline{0.739}_{\pm 0.053}$ & $\textbf{0.211}_{\pm 0.032}$ \\

 & DAPS & $28.49_{\pm 1.52}$ & $0.762_{\pm 0.031}$ & $0.191_{\pm 0.031}$ & $27.25_{\pm 3.18}$ & $0.717_{\pm 0.089}$ & $\underline{0.213}_{\pm 0.055}$ \\
 & DPS & $23.39_{\pm 2.01}$ & $0.623_{\pm 0.082}$ & $0.278_{\pm 0.060}$ & $22.49_{\pm 3.20}$ & $0.591_{\pm 0.101}$ & $0.306_{\pm 0.081}$ \\
 & RED-Diff & $\textbf{30.72}_{\pm 0.56}$ & $\underline{0.793}_{\pm 0.029}$ & $\textbf{0.162}_{\pm 0.032}$ & $\textbf{29.91}_{\pm 1.36}$ & $\textbf{0.751}_{\pm 0.024}$ & $0.214_{\pm 0.080}$ \\

\midrule

\multirow{4}{*}{High dynamic range} 
 & PC-DAPS & $\textbf{28.25}_{\pm 1.17}$ & $\textbf{0.839}_{\pm 0.020}$ & $\textbf{0.163}_{\pm 0.039}$ & $\textbf{26.01}_{\pm 2.77}$ & $\textbf{0.822}_{\pm 0.099}$ & $\textbf{0.173}_{\pm 0.094}$ \\

 & DAPS & $\underline{27.67}_{\pm 3.57}$ & $\textbf{0.839}_{\pm 0.078}$ & $\underline{0.165}_{\pm 0.069}$ & $\underline{25.80}_{\pm 4.37}$ & $\underline{0.819}_{\pm 0.113}$ & $\underline{0.177}_{\pm 0.106}$ \\
 & DPS & $22.73_{\pm 6.07}$ & $\underline{0.591}_{\pm 0.141}$ & $0.264_{\pm 0.156}$ & $19.23_{\pm 2.52}$ & $0.582_{\pm 0.082}$ & $0.503_{\pm 0.106}$ \\
 & RED-Diff & $21.94_{\pm 3.52}$ & $0.508_{\pm 0.081}$ & $0.261_{\pm 0.091}$ & $21.88_{\pm 5.77}$ & $0.598_{\pm 0.097}$ & $0.281_{\pm 0.191}$ \\

\bottomrule

\end{tabular}
}
\end{table*}

%% file: tables/phase-retrieval-outcomes.tex
\centering
\begingroup
\captionof{table}{\textbf{Phase-retrieval outcome frequencies for DAPS and PC-DAPS on FFHQ and ImageNet.} Individual run outcomes classified as Faithful (correctly reconstructed), Upside-down/color-inverted (converged but with rotation or color ambiguities), or Failed (diverged or noisy output). Unrecoverable denotes instances where all 4 runs or particles within a bag are unfaithful, yielding no reliable sample. Overall, PC-DAPS substantially reduces upside-down/color-inverted samples while increasing faithful reconstructions. The performance gap between datasets highlights the inherent difficulty of phase retrieval on highly diverse, unaligned images (ImageNet) compared to more structured prior domains (FFHQ).}
\label{tab:phase-retrieval-outcomes}
\setlength{\tabcolsep}{4pt}
\renewcommand{\arraystretch}{1.35}
\resizebox{\linewidth}{!}{%
    \begin{tabular}{llcc}
        \toprule
        \textbf{Dataset} & \textbf{Outcome} &
        \textbf{DAPS (4 runs)} & \textbf{PC-DAPS (bag of 4)} \\
        \midrule
        \multirow{4}{*}{FFHQ}
        & Faithful $\uparrow$                    & 79.5\% & \textbf{97.5\%} \\
        & Upside-down / color-inverted $\downarrow$ & 14.5\% & \textbf{1.5\%} \\
        & Failed $\downarrow$            & 6.0\%  & \textbf{1.0\%} \\
        & Unrecoverable runs/bags $\downarrow$            & 4\% & \textbf{1\%} \\
        \midrule
        \multirow{4}{*}{ImageNet}
        & Faithful $\uparrow$                    & 38.5\% & \textbf{45.0\%} \\
        & Upside-down / color-inverted $\downarrow$ & 42.0\% & \textbf{30.5\%} \\
        & Failed $\downarrow$            & \textbf{19.5\%} & 24.5\% \\
        & Unrecoverable runs/bags $\downarrow$            & 25\% & \textbf{15\%} \\
        \bottomrule
    \end{tabular}%
}
\endgroup

%% file: sections/6-discussion.tex
\section{Discussion}
\label{sec:discussion}

\paragraph{Motivation, limitations, and future work.} Results in the diffusion literature for ambiguous inverse problems are usually reported as best-of-$k$, with selection using the ground truth. This evaluation approach became common because existing solvers are inconsistent across runs, and a single-sample result would have large variations in quality. We see this instability as a symptom of premature mode selection in the context of measurement ambiguity. We introduce PC-DAPS to utilize the learned prior information to reduce this unwanted variance during sampling without the need for a ground-truth reference. Even in high-dimensional datasets, including FFHQ, our method demonstrates significant improvements due to the local structure of the prior. PC-DAPS obtains a smaller improvement on more structurally diverse datasets, namely ImageNet, because the prior landscape is more complex, exposing a remaining limitation and open problem to be addressed. We encourage future work to further explore our proposed paradigm of leveraging the prior score to assess candidate reconstructions at inference time, and work towards eliminating prior-inconsistent reconstructions even under less-structured priors.

\paragraph{Posterior sampling metrics.} Standard metrics like PSNR, SSIM, and LPIPS measure how similar a generated image is to a single reference image. Observe that these evaluation metrics penalize methods that find several plausible solutions that are equally valid under the measurement and prior if the solutions are dissimilar from the fixed reference. Because sample diversity and likelihood consistency are explicit goals of posterior sampling, including in applications such as uncertainty quantification, these sample-quality metrics are not adequate replacements for quality of the sampled posterior. In our work, we report a separate axis of reconstruction quality: prior consistency. We believe this or similar multifaceted metrics should be considered alongside standard fidelity metrics. 

%% file: sections/99-acknowledgements.tex
\subsubsection*{Acknowledgments}
MK is supported by the German Federal Ministry of Education and Research (BMBF) through the DAAD PROMOS program and by the Konrad Zuse School of Excellence in Learning and Intelligent Systems (ELIZA). BAB is supported by the National Science Foundation Graduate Research Fellowship Program under Grant No. DGE-2039655. Any opinions, findings, and conclusions or recommendations expressed in this material are those of the author(s) and do not necessarily reflect the views of the National Science Foundation. AG is a member of the Machine Learning Cluster of Excellence, EXC 2064/1, project number 390727645.

%% file: sections/A-appendix.tex
\section{Supplementary material}

\subsection{Full algorithm}
\label[appendix]{app:algorithm}

\begin{algorithm}[H]
\caption{Prior-Consistent DAPS (PC-DAPS)}
\label{alg:pcdaps}
\begin{algorithmic}[1]
\Require measurement $y$, pretrained denoiser $D_\theta$, measurement operator $\mathcal{A}$, measurement noise $\beta_y$
\Require noise schedule $\sigma_{t_{N_A}} > \dots > \sigma_{t_0}=0$, bag size $K$, Langevin steps $N$, threshold $\gamma$
\State $x_{t_{N_A}}^{(k)} \sim \mathcal{N}(0,\sigma_{t_{N_A}}^2 I)$ for $k=1,\dots,K$
\State $\texttt{pruned} \gets \texttt{false}$
\For{$i=N_A,N_A-1,\dots,1$}

    \Statex \hspace{0.5em}\emph{Clean-signal estimates and denoising distribution}
    \State $\tilde{x}_0^{(k)} \gets
    D_\theta\!\left(x_{t_i}^{(k)};\sigma_{t_i}\right)$
    \If{$\texttt{pruned}=\texttt{false}$}
        \State $q(x_0\mid x_t)\gets
        \frac{1}{K}\sum_{k=1}^K
        \mathcal{N}\!\left(x_0;\tilde{x}_0^{(k)},
        \operatorname{diag}(\Var_{\mathcal{B}}(\tilde{x}_0^{(k)}))\right)$
    \Else
        \State cluster $\{\tilde{x}_0^{(k)}\}_{k=1}^K$ into $\{\mathcal{C}_m\}_{m=1}^M$
        \State construct $q(x_0\mid x_t)$ using
        \cref{eq:pcdaps-mixture,eq:pcdaps-mixture-params}
    \EndIf

    \Statex \hspace{0.5em}\emph{Langevin refinement}
    \If{$i>1$}
        \State $\hat{x}_0^{(k,0)}\gets\tilde{x}_0^{(k)}$
        \For{$j=0,\dots,N-1$}
            \State $\hat{x}_0^{(k,j+1)}\gets$
            Langevin step (\cref{eq:daps-langevin})
            using the score in \cref{eq:pcdaps-mixture-score}
        \EndFor

        \Statex \hspace{0.5em}\emph{One-shot consensus pruning}
        \If{$\neg\texttt{pruned}$ and
            $\bar r < r^\star$ and clustering returns $M>1$ clusters}
            \State compute $s(\mathcal{C}_m)$ for each cluster using
            \cref{eq:pcdaps-prior-score}
            \State $\mathcal{G}\gets
            \{\mathcal{C}_m:s(\mathcal{C}_m) \geq (1+\gamma) \max s(\mathcal{C}_m)\}$
            \State randomly replace particles in clusters
            $\mathcal{C}_m\notin\mathcal{G}$ by copies from clusters in $\mathcal{G}$
            \State $\texttt{pruned}\gets\texttt{true}$
        \EndIf

        \Statex \hspace{0.5em}\emph{Re-noising}
        \State $x_{t_{i-1}}^{(k)}\gets
        \hat{x}_0^{(k,N)}+\sigma_{t_{i-1}}\epsilon^{(k)}$,
        \quad $\epsilon^{(k)}\sim\mathcal{N}(0,I)$
    \Else
        \State $\hat{x}_0^{(k)}\gets
        \textsc{ODE-Solve}(x_{t_1}^{(k)},\sigma_{t_1}\to0)$
    \EndIf

\EndFor
\State \Return $\{\hat{x}_0^{(k)}\}_{k=1}^K$
\end{algorithmic}
\end{algorithm}

\subsection{Toy experiment details}
\label[appendix]{app:toy-experiment}
\paragraph{Prior.}
The signals $x$ are elements of $\mathbb{R}^2$. We define a reference vector $u$, in this case the first standard basis vector $u = e_1$. The prior has two atoms at $\pm a u$ for scalar $a > 0$:
\begin{equation}
    p(x_0) = \tfrac{1}{2}\,\delta_{-a u} + \tfrac{1}{2}\,\delta_{a u}, \quad a = 1 .
\end{equation}
Observe that the prior denoiser is known exactly for EDM-parametrized forward model $x_\sigma = x_0 + \sigma \epsilon$ with $\epsilon \sim \mathcal{N}(0, I)$:
\begin{equation}
    D(x, \sigma) = \mathbb{E}[x_0 \mid x_\sigma = x] = a \tanh\!\big(a \langle x, u \rangle / \sigma^2\big)\, u .
\end{equation}
DAPS and PC-DAPS use this closed-form denoiser in place of a trained network, both for the probability-flow ODE and for all Tweedie estimates.

\paragraph{Measurement and likelihood.}
The measurement operator is a quadratic function in the inner product:
\begin{equation}
    \mathcal{A}(x) = f(\langle x, u \rangle), \quad f(z) = \frac{c}{2ab}\,(z - a)(z + a - b),
\end{equation}
chosen to interpolate $f(-a) = c$, $f(a+b) = c$ and $f(a) = 0$. Measurements follow $y = \mathcal{A}(x_0) + n$ with $n \sim \mathcal{N}(0, \beta_y^2)$, so the likelihood is $p(y \mid x_0) = \mathcal{N}(y; \mathcal{A}(x_0), \beta_y^2)$. The observation is set to $y = c = \kappa \beta_y$. The precise values used in \cref{fig:toy-example} are $\beta_y = 0.05$, $b = 0.1$, $\kappa = 2$ (thus $c = 0.1$).

\paragraph{Exact posterior.} By Bayes' rule, the ground truth posterior $p(x_0 \mid y)$ is accessible in closed form:
\begin{equation}
    p(x_0 \mid y) = (1 - \pi)\,\delta_{-a u} + \pi\,\delta_{a u}, \quad \pi
    =
    \frac{p(y \mid a u)}
    {p(y \mid -a u) + p(y \mid a u)}.
\end{equation}
Observe that although $\mathcal{A}((a+b)u) = \mathcal{A}(-au)$, $(a+b)u$ has no prior mass, thus samples from (the neighborhood of) $(a+b)u$ are prior-inconsistent hallucinations.

\subsection{More details on Gaussian mixture approximation}
\label[appendix]{app:mixture}

To illustrate the role of the Gaussian mixture $q(x_0 \mid x_t)$ introduced in \cref{sec:method-mixture}, used by PC-DAPS to approximate the denoising distribution $p(x_0 \mid x_t)$, we show a side-by-side comparison of the Langevin dynamics refinement when using (i) the DAPS Gaussian approximation $\mathcal{N}(x_0; \hat x_0(x_t), r_t^2 I)$ and (ii) the PC-DAPS GMM approximation $q(x_0 \mid x_t)$ on a toy example.

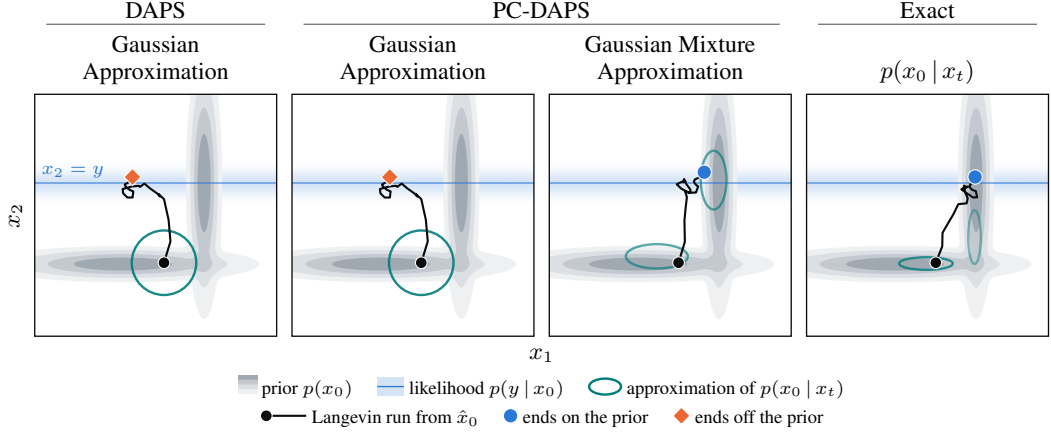
\begin{figure*}[t]
    \centering
    \begingroup
    \def\lvdir{figures/gmm-toy-exp/}
    \input{figures/gmm-toy-exp/gmm-toy-exp}
    \caption{\textbf{Gaussian mixture approximations preserve prior-supported modes in Langevin refinement.} Each pane illustrates $N$ iterations of Langevin refinement from the same starting point $\hat{x}_0$ (black dot; path smoothed for readability). The same prior $p(x_0)$ (gray) and likelihood $p(y\mid x_0)$ (blue) are shown in each pane, while the approximation $q(x_0\mid x_t)$ (teal) differs. When utilizing the Gaussian approximation, both DAPS (left) and PC-DAPS (middle-left) produce Langevin targets which are measurement-consistent yet prior-inconsistent, and the refinement ends outside the prior (orange diamond). In contrast, the PC-DAPS Gaussian mixture (middle-right) represents multiple prior-supported regions explored by particles in the bag, allowing Langevin refinement to move toward the prior-supported mode associated with the measurement (blue dot). The exact denoising distribution (right) exhibits the same multimodal structure.}
    \label{fig:gmm-toy}
    \endgroup
\end{figure*}

In the illustrated example depicted in \cref{fig:gmm-toy}, we consider a multimodal prior distribution on the plane $\mathbb{R}^2$, and construct $\mathcal{A}$ to simply select the second component $x_2$ of the signal vector $x$. Four panes are shown depicting the Langevin algorithm procedure of a single (i) DAPS iteration, (ii) PC-DAPS iteration using the DAPS Gaussian approximation, (iii) PC-DAPS iteration using the PC-DAPS GMM approximation, and (iv) ground truth (denoiser and Langevin) iteration. The high-probability level sets of both the (approximated) denoiser and the induced Langevin target distribution are depicted in each pane.

Observe that, in both panes using the DAPS Gaussian approximation, the Langevin algorithm effectively searches for points (i) consistent with the provided measurement $y = x_2$ and (ii) close to the initialization point in a Euclidean sense. As a result, the algorithm terminates by returning a likelihood-consistent yet posterior-inconsistent point. In contrast, the Gaussian mixture approximation is attracted to the region of the measurement's preimage that is consistent with the prior thanks to other particles in the bag which explored and identified prior-consistent regions distant from the Langevin initialization. Additionally, observe that although the DAPS Gaussian is isotropic and thus places high probability outside the high-probability level set of the prior, the PC-DAPS approximation's component covariances better adapt to the geometry of the prior's support.

\subsection{Additional implementation choices}
\label[appendix]{app:implementation}

There are two implementation choices that are orthogonal to the core PC-DAPS algorithm but improve reconstruction quality or computational efficiency in practice: skipping the final Langevin refinement and using adaptive ODE stepping. Their effects are summarized in \cref{tab:implementation-ablation} and described in detail below.

\paragraph{Skipping the last Langevin refinement.} At the final annealing time ($t_0 = 0$), the Langevin update is not performed, and the ODE endpoint $\hat{x}_0(x_{t_0})$ is used as the output. When the noise is nearly zero, the Langevin step size becomes very small, and the added noise can slightly reduce sample quality.

\paragraph{Adaptive ODE steps.} To compute the DAPS clean estimate $\hat{x}_0(x_{t_i})$, the probability flow ODE is integrated using Euler steps. One Euler step is used at high noise levels, where the clean estimate is less sensitive to the precise ODE integration, and this is increased to five steps at lower noise levels, where the prior becomes more discriminative and finer integration reduces discretization error near the data manifold. Since PC-DAPS utilizes a bag of multiple particles, reducing unnecessary ODE evaluations is important for controlling the additional computational cost. Adaptive ODE stepping is therefore used to improve the time efficiency of the particle-based sampler.

Descriptions of the implementation details of the particle clustering and pruning procedure used by PC-DAPS are given below.

\paragraph{Clustering technique.} To assign particles to components, the corresponding clean samples $\{\tilde x_0^{(k)}\}_{k=1}^K$ described in \cref{sec:method-mixture} are downsampled to a coarse resolution (e.g., $32{\times}32$), centered, and projected onto the leading principal components that explain $90\%$ of the variance. On this subspace, a Gaussian mixture model is fit with $M$ selected via Bayesian information criterion \citep{schwarz1978estimating}. Each particle is then hard-assigned to its most likely component, giving a partition $\{\mathcal{C}_m\}_{m=1}^{M}$ of the bag.

\paragraph{Pruning threshold.} Pruning is applied to the post-Langevin samples only when the test in \cref{eq:pcdaps-jensen-quantities} fires ($r^\star > \bar r$) and the clustering returns more than one cluster. Each cluster is scored using the non-positive function \cref{eq:pcdaps-prior-score} with $\mathcal{S}=\{0.5,1,2\}$. Denoting $s^\star=\max_m s(\mathcal{C}_m)$, all clusters $\mathcal{C}_m$ satisfying $s(\mathcal{C}_m) < (1+\gamma)s^\star$ are pruned, while clusters satisfying $ (1+\gamma)s^\star \leq s(\mathcal{C}_m) \leq s^\star$ are preserved. Each pruned particle is replaced by a copy of a surviving particle, cycling through the survivors. Observe that the cluster producing score $s^\star$ always survives. $\gamma$ is calibrated from within-cluster score variation: following \citet{zhang2025improvingdiffusioninverseproblem}, 10 calibration images per dataset are denoised via PC-DAPS (including the clustering procedure) without pruning. The relative average score difference for all particle pairs within each cluster is computed, and $\gamma$ is set to the 95th percentile of these differences.

\input{tables/implementation-ablation}

Ablations of the main computational parameters and their comparisons are presented in the following paragraphs.

\begin{figure*}[t]
\centering
\includegraphics[width=\textwidth]{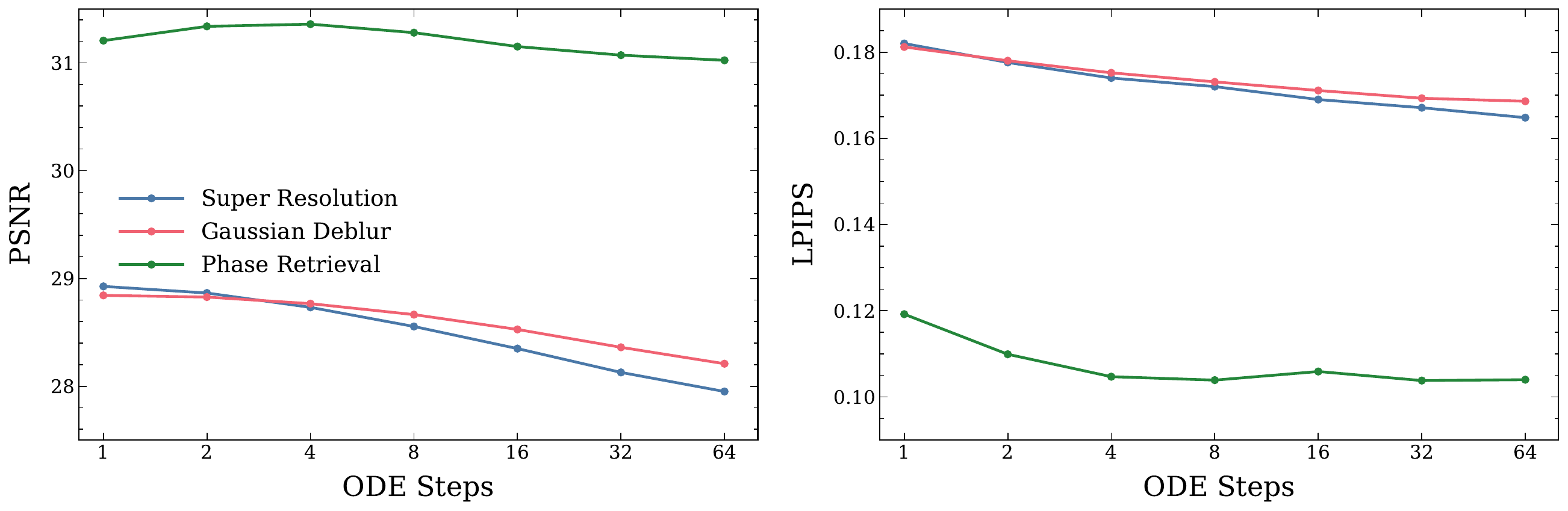}
\caption{\textbf{Effect of the number of ODE steps on PC-DAPS.} Increasing the number of ODE steps generally improves perceptual quality (LPIPS) but can reduce PSNR, with the effect depending on the inverse problem.}
\label{fig:ode-steps}
\end{figure*}

\begin{figure*}[t]
\centering
\includegraphics[width=\textwidth]{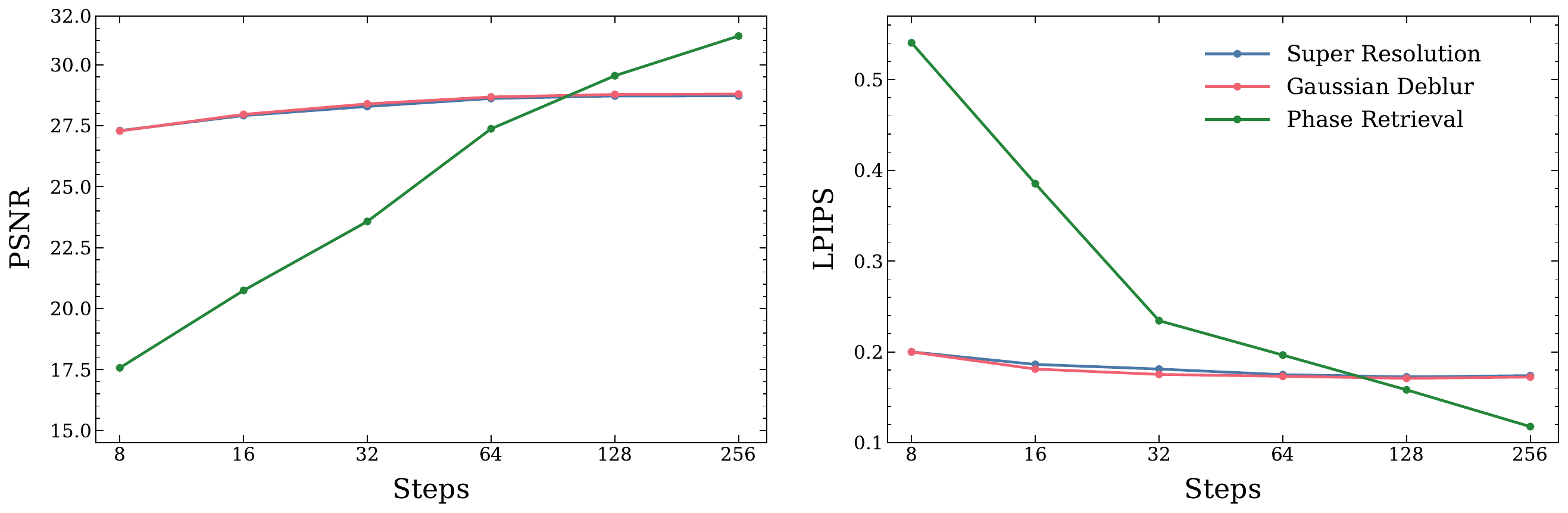}
\caption{\textbf{Effect of the number of annealing noise scheduling steps on PC-DAPS.} Increasing the number of annealing steps has little effect on super-resolution and Gaussian deblurring, while phase retrieval shows substantial improvement with a finer annealing schedule.}
\label{fig:annealing-steps}
\end{figure*}

\paragraph{Ablation of ODE steps and annealing noise scheduling steps.} PC-DAPS benefits from the same implementation improvements identified in DAPS. The effects of the number of ODE steps used to compute the clean estimate and the number of annealing noise-scheduling steps are ablated, and trends consistent with those reported for DAPS are observed: increasing the number of ODE steps generally improves perceptual quality, while denser annealing schedules are particularly beneficial for nonlinear inverse problems. The corresponding ablations are shown in \cref{fig:ode-steps} and \cref{fig:annealing-steps}.

\begin{figure*}[t]
\centering
\includegraphics[width=\textwidth]{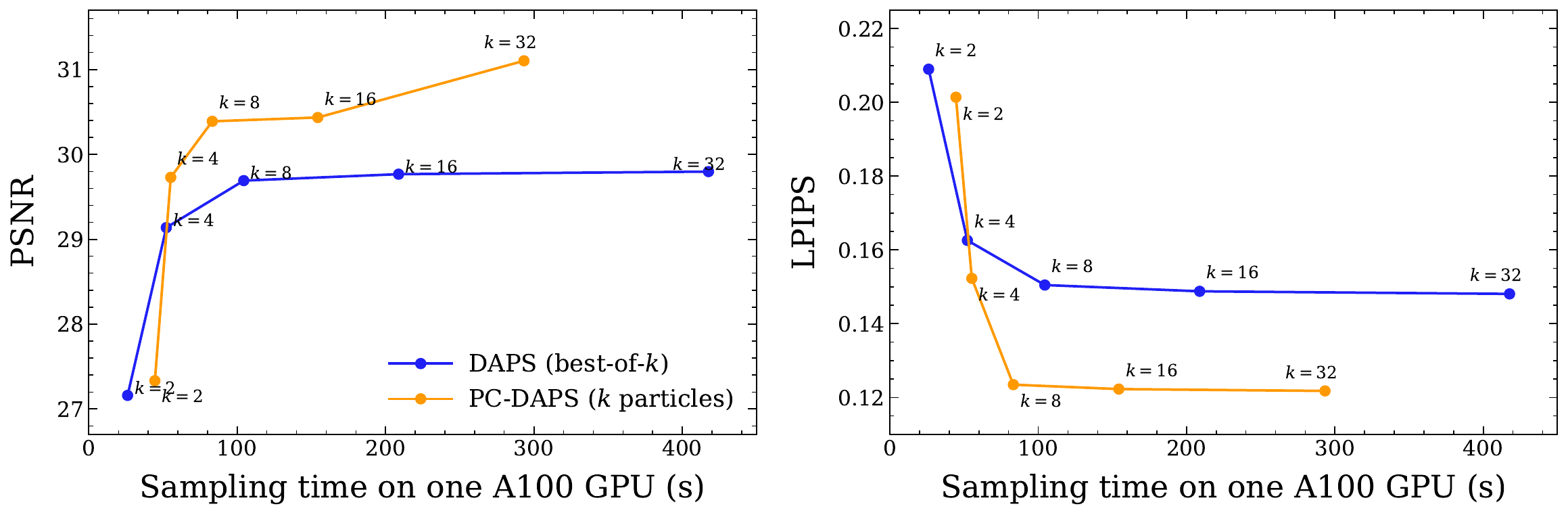}
\caption{\textbf{Effect of the number of particles in a bag on phase retrieval.}
Increasing the bag size improves PC-DAPS reconstruction quality, with both PSNR and LPIPS
improving rapidly up to $K=8$--$16$ and then beginning to saturate. PC-DAPS is competitive with DAPS best-of-$k$ at comparable values of $k$, while requiring less sampling time
for larger $k$.}
\label{fig:bag-size}
\end{figure*}

\paragraph{Number of particles in a bag.} Using more particles provides the selection procedure with more candidate trajectories, which improves its ability to distinguish distinct modes and reject prior-inconsistent but measurement-consistent components. On phase retrieval, increasing the bag size substantially improves both PSNR and LPIPS, with the gains becoming smaller beyond $K=8$--$16$. This improvement comes at an approximately linear increase in computational cost, since all $K$ particles are propagated through the annealing schedule. Compared with DAPS best-of-$k$, PC-DAPS achieves higher reconstruction quality while using less sampling time for larger bag sizes, as shown in \cref{fig:bag-size}.

\paragraph{A fair comparison on phase retrieval.} \Cref{tab:main-results} reports the best-of-$k$ results used by the baseline methods, while PC-DAPS is evaluated using the mean over its $K$ samples. For a direct comparison under the same evaluation protocol, \cref{tab:phase-retrieval-mean-of-k} reports mean-of-$k$ results for all methods. The baseline methods show lower mean performance because their $k$ reconstructions can include measurement-consistent but prior-inconsistent solutions, such as color inversions, rotations, or translations, whereas best-of-$k$ selects only the reconstruction closest to the ground truth.

\input{tables/phase-retrieval-mean-of-k}

\subsection{Time cost}
\label[appendix]{app:time-cost}

PC-DAPS keeps $K$ particles in one sampling trajectory, while DAPS needs $K$ separate runs to get $K$ candidate reconstructions. If implementation overhead is ignored, both methods need about the same number of denoiser evaluations, which increases roughly linearly with $K$. PC-DAPS does have extra steps, including clustering, estimating mixture parameters, and a one-shot pruning step. However, these add only a small overhead compared to the cost of the ODE and Langevin updates.

\begin{figure*}[t]
\centering
\includegraphics[width=\textwidth]{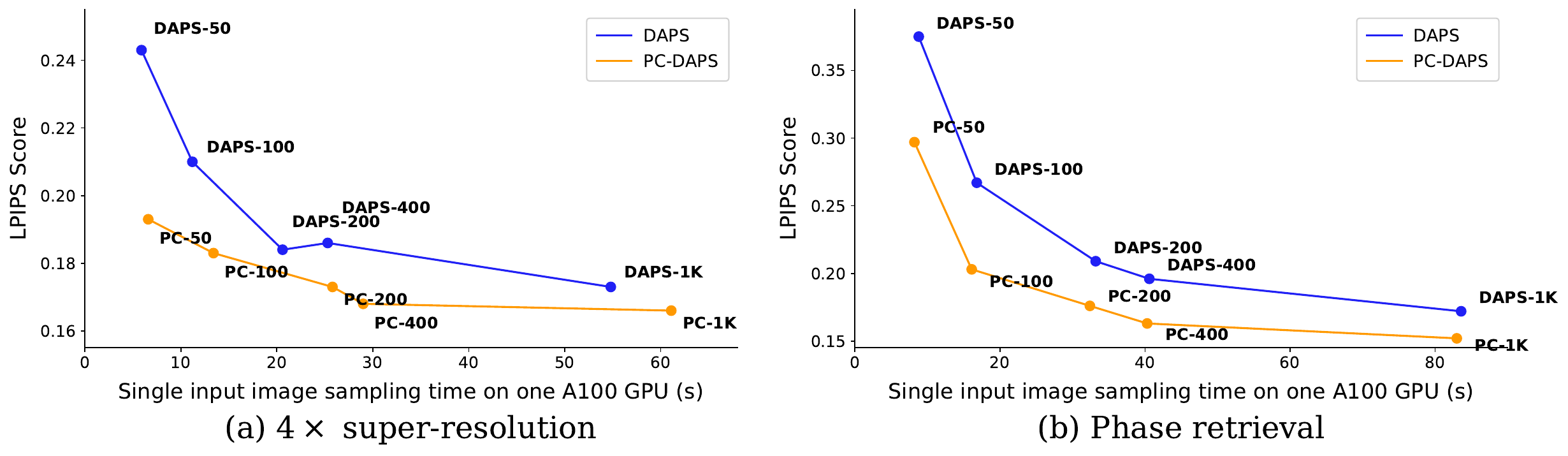}
\caption{\textbf{Comparison of single input sampling time and reconstruction quality between DAPS and PC-DAPS.} The $x$-axis shows the sampling time for a single input sample measured on an NVIDIA A100-SXM4-80GB GPU, while the $y$-axis reports LPIPS. Evaluation is performed on 100 FFHQ images. For phase retrieval, DAPS requires four independent runs for an input image; therefore, the reported DAPS sampling time corresponds to the total time of all four runs.}
\label{fig:sampling-time}
\end{figure*}

The implementation choices in \cref{app:implementation}, such as skipping the final Langevin refinement and using adaptive ODE steps at low noise, help lower the sampling cost of PC-DAPS. \Cref{fig:sampling-time} shows the measured wall-clock time and reconstruction quality. For super-resolution, PC-DAPS takes longer than DAPS because it uses multiple particles. For phase retrieval, though, the DAPS runtime includes four separate runs because it uses a best-of-4 evaluation protocol. PC-DAPS, on the other hand, creates the particle set in one run and selects internally. These results suggest that using multiple particles does not always mean a much higher wall-clock cost to achieve the same quality metric.

\subsection{Inverse problem setup}
\label[appendix]{app:inverse-setup}

In \cref{tab:main-results}, following the inverse-problem settings of DAPS \citep{zhang2025improvingdiffusioninverseproblem}, the benchmark includes the eight tasks from DAPS: $4{\times}$ super-resolution, box inpainting, random inpainting, Gaussian deblurring, motion deblurring, nonlinear deblurring, phase retrieval, and high dynamic range (HDR) reconstruction. For these tasks, the same measurement operators, measurement noise level, and evaluation protocol as DAPS are used. Unless otherwise stated, measurements are corrupted with additive Gaussian noise with standard deviation $\beta_y = 0.05$. Methods are evaluated on 100 test images from each dataset, FFHQ and ImageNet, at a resolution of $256{\times}256$. Additionally, $16{\times}$ super-resolution and high-pass filtering from \citet{thaker2025frequency} are included in the table and described in detail below.

\paragraph{Super-resolution $16{\times}$.}
Bicubic downsampling is applied from $256{\times}256$ to $16{\times}16$, followed by additive Gaussian noise with standard deviation $\beta_y = 0.05$. This uses the same linear bicubic downsampling operator as the $4{\times}$ super-resolution task in DAPS, but with a larger downsampling factor. The resulting operator has a larger null space, leaving more image content unconstrained by the measurement and providing a more challenging setting. All other sampler settings follow the $4{\times}$ super-resolution configuration.

\paragraph{High-pass filtering.}
Following \citet{thaker2025frequency}, a $61{\times}61$ convolutional kernel is used, constructed as a Dirac delta minus a Gaussian blur kernel with standard deviation $5.0$. The resulting kernel has zero sum and therefore removes the DC component. Gaussian noise with standard deviation $\beta_y = 0.05$ is added to the filtered image.

\subsubsection{Baseline details}
\label[appendix]{app:baseline-details}

For DAPS \citep{zhang2025improvingdiffusioninverseproblem}, DPS \citep{chung2023diffusion}, DDNM \citep{wang2023zeroshot}, FPS-SMC \citep{dou2024diffusion}, RED-Diff \citep{mardani2024variational}, FGPS \citep{thaker2025frequency}, PnP-DM \citep{wu2024principled}, and DPnP \citep{xu2024provably}, the authors' official public implementations and recommended task-specific hyperparameters are used throughout the benchmark. DAPS uses the DAPS-4K configuration for nonlinear tasks and DAPS-1K for linear tasks. For consistency in computational budget, PC-DAPS uses the DAPS-1K configuration for both linear and nonlinear tasks, including its adaptive variant across all tasks. When a baseline specifies task-specific step sizes or noise-schedule parameters, the reported values are used without modification. HIO, a classical phase retrieval method, is included as an additional reference baseline, with results taken from \citet{zhang2025improvingdiffusioninverseproblem}.

For the two additional tasks not included in the DAPS benchmark, $16{\times}$ super-resolution and high-pass filtering, the closest corresponding configuration available for each baseline is used: the $4{\times}$ super-resolution configuration for $16{\times}$ super-resolution and the Gaussian deblurring configuration for high-pass filtering. Only the measurement operator and the associated measurement generation are modified to match the task definitions described above.

Unless otherwise specified, PC-DAPS uses the $1$K setting with 200 annealing steps and adaptive ODE stepping, and generates $K=4$ reconstructions for each measurement. Reported PC-DAPS metrics are averaged over the four reconstructions. For phase retrieval, baselines other than PC-DAPS are evaluated using the best of four independent runs, selecting the reconstruction closest to the ground-truth image, following prior work \citep{chung2023diffusion,zhang2025improvingdiffusioninverseproblem}. This protocol gives the baselines access to ground-truth information that is unavailable to PC-DAPS at test time. PSNR, SSIM \citep{wang2004image}, and LPIPS \citep{zhang2018unreasonable} are reported in \cref{tab:main-results} for all methods. For ablation studies, the PC-DAPS variants follow the corresponding DAPS settings reported in \cref{tab:sampling-configurations}.

\input{tables/sampling-configurations}
\newpage

\subsection{More qualitative samples}
\label[appendix]{app:more-samples}

\begin{figure}[H]
\centering
\includegraphics[width=0.9\textwidth]{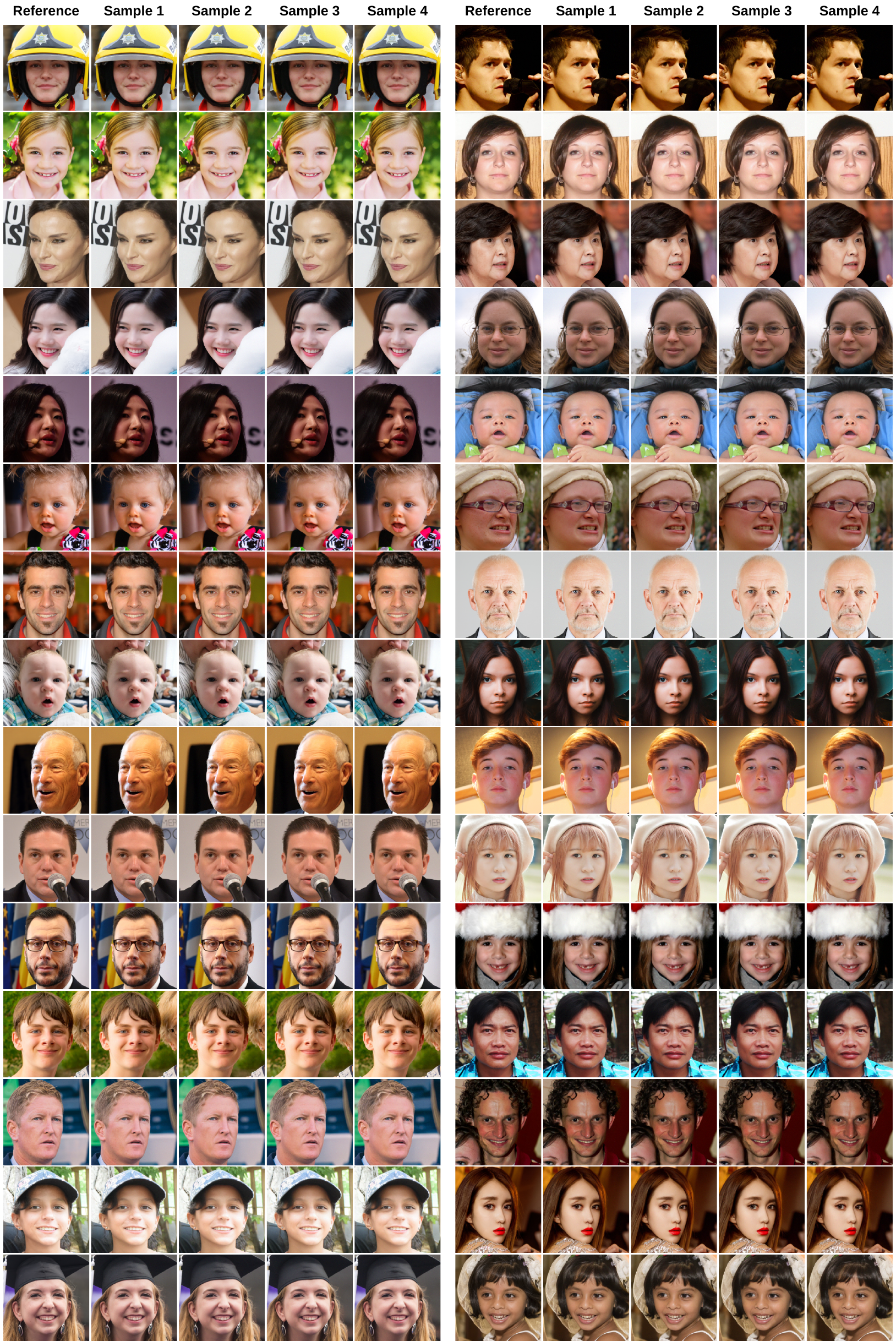}
\caption{\textbf{PC-DAPS samples for phase retrieval on FFHQ.} Samples are shown without manual selection to better visualize success rate and sample quality.}
\label{fig:samples-phase-retrieval-ffhq}
\end{figure}

\begin{figure}[t]
\centering
\includegraphics[width=0.9\textwidth]{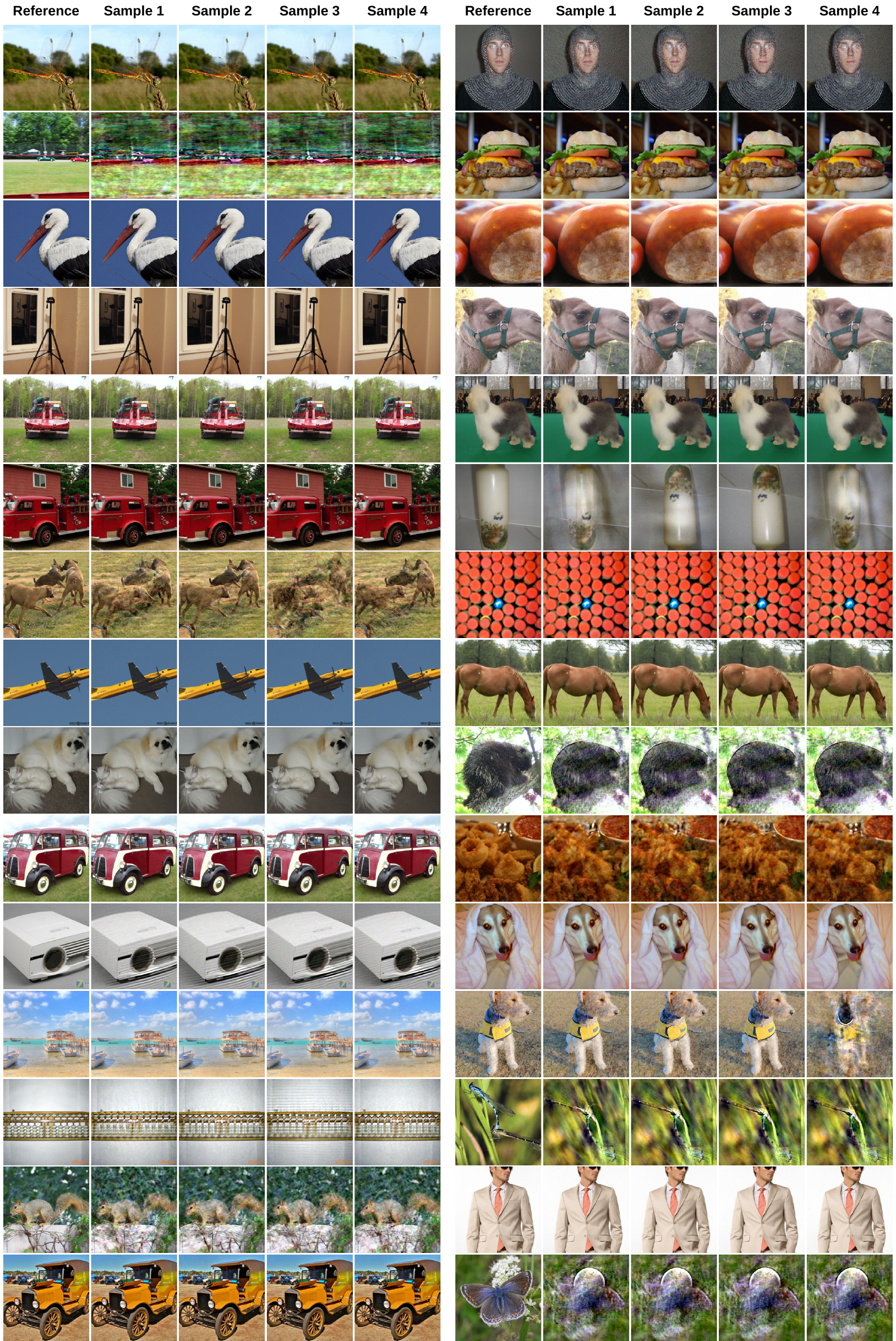}
\caption{\textbf{PC-DAPS samples for phase retrieval on ImageNet.} Samples are shown without manual selection to better visualize the success rate and sample quality. As discussed in \cref{sec:results-hallucinations}, ImageNet samples in a bag still contain failed (diverged/noisy) particles and underperform compared to FFHQ.}
\label{fig:samples-phase-retrieval-imagenet}
\end{figure}

\begin{figure}[t]
\centering
\includegraphics[width=0.9\textwidth]{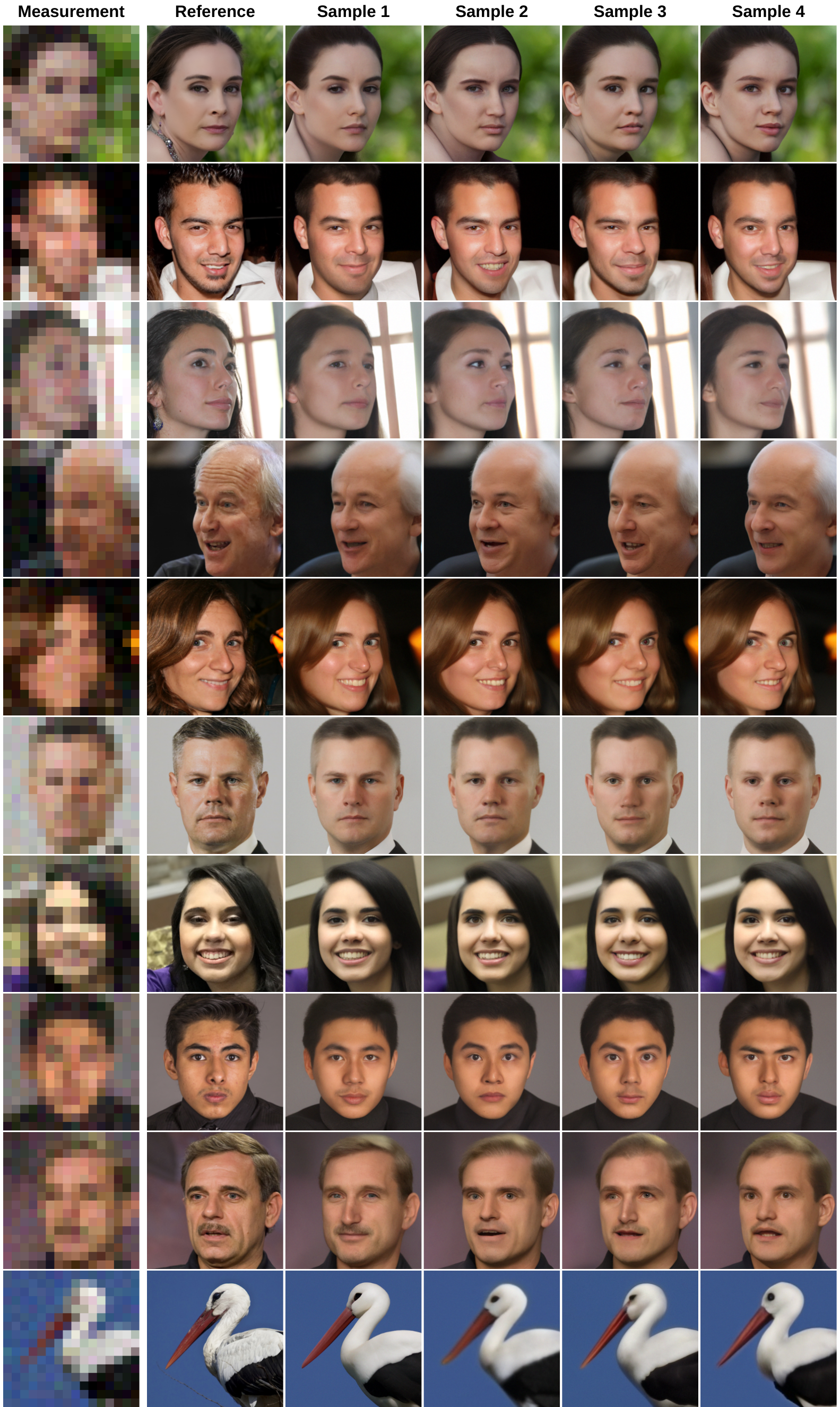}
\caption{\textbf{PC-DAPS samples for $16\times$ super-resolution.} PC-DAPS is able to keep the diversity in the multimodal posterior distribution.}
\label{fig:samples-super-resolution}
\end{figure}

\begin{figure}[t]
\centering
\includegraphics[width=0.9\textwidth]{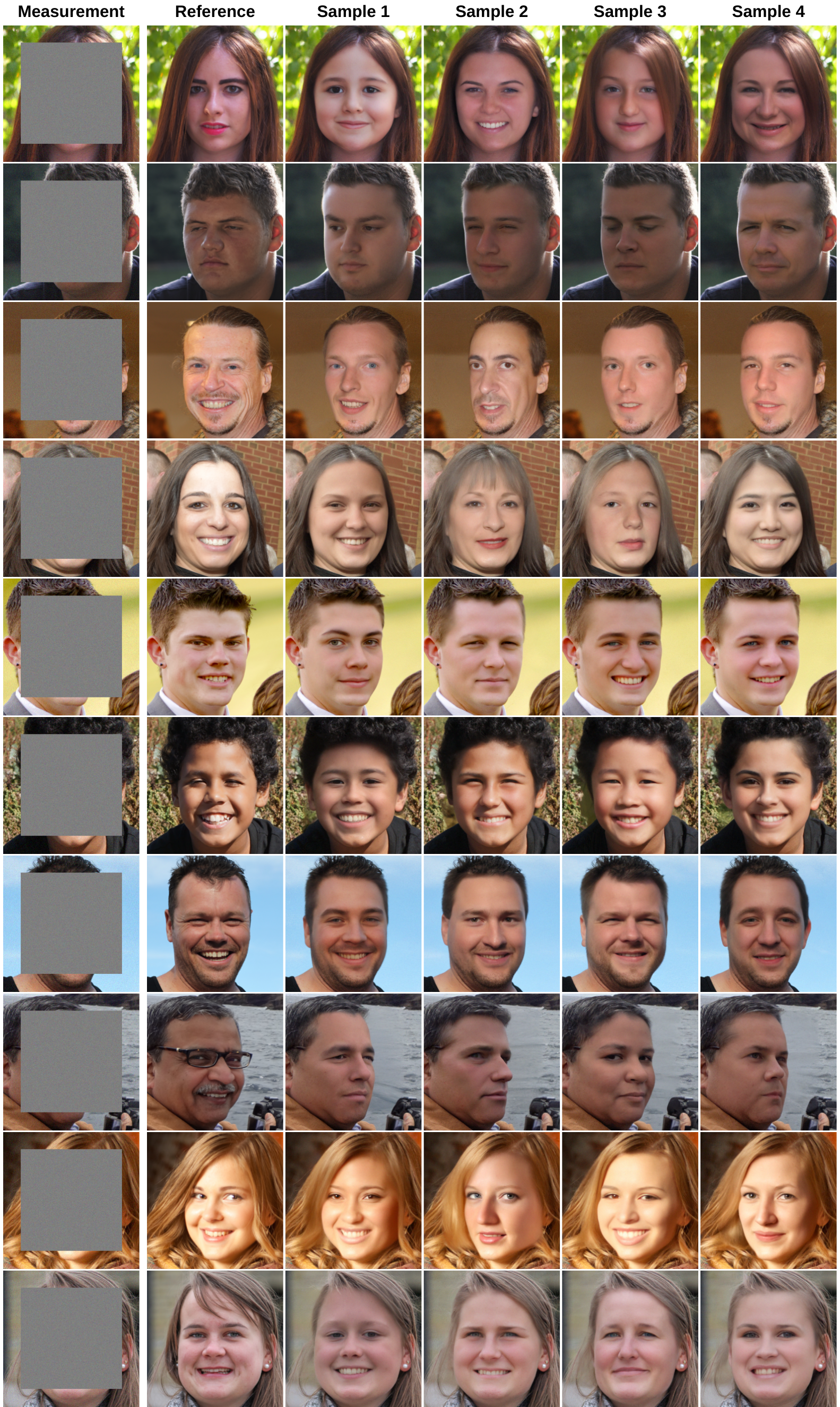}
\caption{\textbf{PC-DAPS samples for $190{\times}190$ box inpainting.} PC-DAPS is able to keep the diversity in the multimodal posterior distribution.}
\label{fig:samples-inpainting}
\end{figure}

\begin{figure}[t]
\centering
\includegraphics[width=0.9\textwidth]{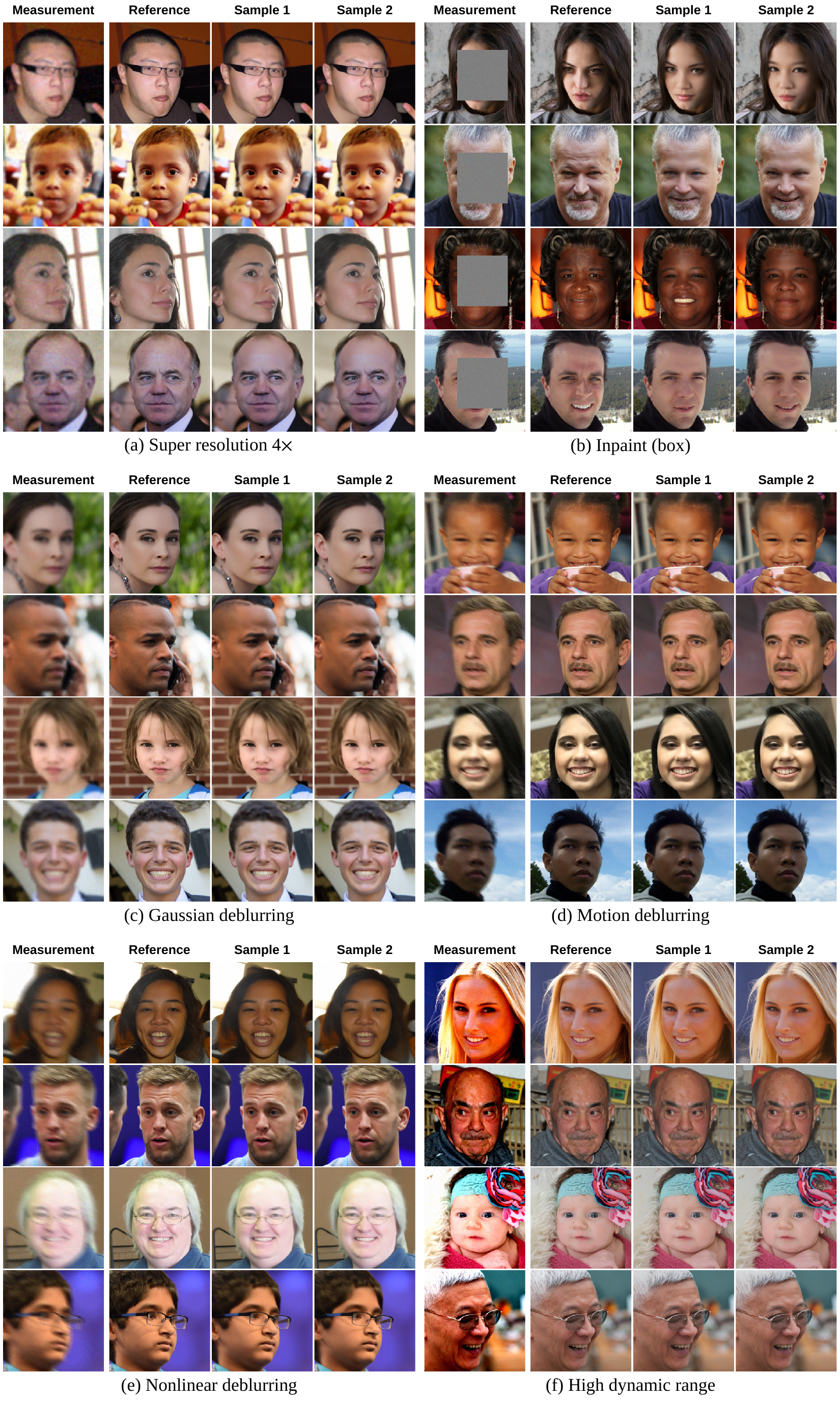}
\caption{\textbf{PC-DAPS samples for various tasks on FFHQ.} PC-DAPS obtains competitive reconstructions for linear and nonlinear tasks.}
\label{fig:samples-ffhq}
\end{figure}

\begin{figure}[t]
\centering
\includegraphics[width=0.9\textwidth]{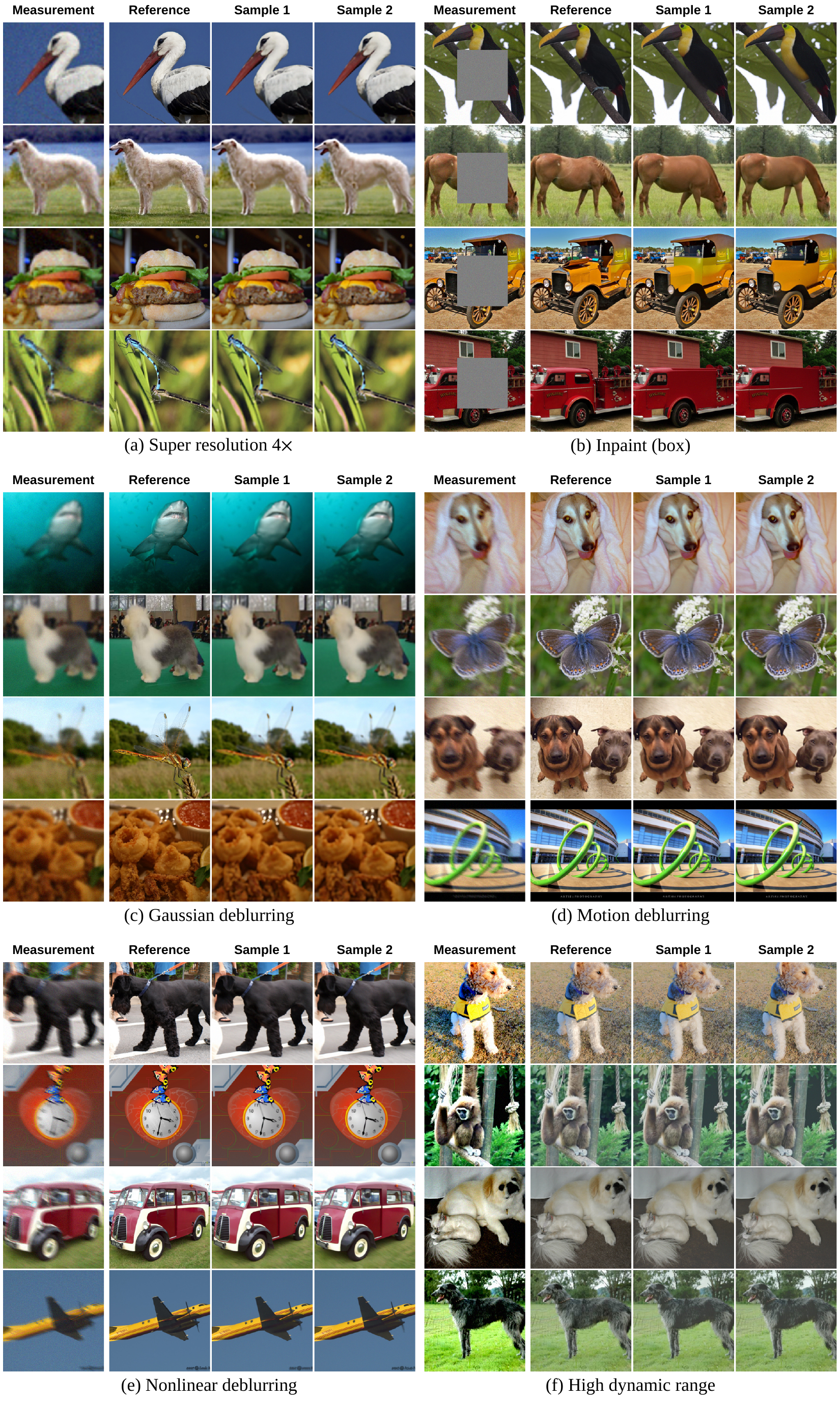}
\caption{\textbf{PC-DAPS samples for various tasks on ImageNet.} PC-DAPS obtains competitive reconstructions for linear and nonlinear tasks.}
\label{fig:samples-imagenet}
\end{figure}

%% file: figures/gmm-toy-exp/gmm-toy-exp.tex
\providecommand{\lvdir}{}%
\definecolor{lvBlue}{HTML}{2A78D6}%
\definecolor{lvLik}{HTML}{2A78D6}%
\definecolor{lvOrange}{HTML}{EB6834}%
\definecolor{lvInk}{HTML}{0B0B0B}%
\definecolor{lvGray}{HTML}{52514E}%
\definecolor{lvPathLo}{HTML}{B9B7AE}%
\definecolor{lvTargetLo}{HTML}{F7CDB7}%
\definecolor{lvTargetHi}{HTML}{EF9670}%
\definecolor{lvTeal}{HTML}{007C7A}%
\tikzset{%
  lvstart/.style={circle, fill=lvInk, draw=white, line width=0.4pt, inner sep=0pt, minimum size=4.4pt},%
  lvend/.style={star, star points=5, star point ratio=2.3, fill=lvInk, draw=white, line width=0.35pt, inner sep=0pt,%
    minimum size=8pt},%
  lvbag/.style={circle, fill=white, draw=lvGray, line width=0.6pt, inner sep=0pt, minimum size=4.2pt},%
  lvgood/.style={circle, fill=lvBlue, draw=white, line width=0.4pt, inner sep=0pt, minimum size=5.4pt},%
  lvbad/.style={diamond, fill=lvOrange, draw=white, line width=0.4pt, inner sep=0pt, minimum size=6.6pt}}%
\ifdefined\lvGlyphPrior\else%
  \newsavebox{\lvGlyphPrior}\newsavebox{\lvGlyphLik}\newsavebox{\lvGlyphApprox}\newsavebox{\lvGlyphPath}%
  \newsavebox{\lvGlyphBag}\newsavebox{\lvGlyphDraws}%
  \newsavebox{\lvGlyphApproxOne}\newsavebox{\lvGlyphDrift}\newsavebox{\lvGlyphEnds}\newsavebox{\lvGlyphTarget}%
  \newsavebox{\lvGlyphGood}\newsavebox{\lvGlyphBad}%
\fi%
\sbox{\lvGlyphPrior}{\includegraphics[width=7pt, height=7pt]{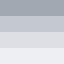}}%
\sbox{\lvGlyphLik}{\tikz[baseline=-0.6ex]{\fill[lvLik, opacity=0.2] (0,-3.5pt) rectangle (10pt,3.5pt);%
  \draw[lvLik, line width=0.5pt] (0,0) -- (10pt,0);}}%
\sbox{\lvGlyphApprox}{\tikz[baseline=-0.6ex]{\draw[lvOrange, line width=0.8pt] (0,0) ellipse [x radius=5pt, y radius=2.6pt];%
  \draw[lvOrange, line width=0.45pt, opacity=0.9] (0,0) ellipse [x radius=7.5pt, y radius=3.9pt];}}%
\sbox{\lvGlyphPath}{\tikz[baseline=-0.6ex]{\draw[lvPathLo, line width=0.6pt] (0,0) -- (7pt,0);%
  \draw[lvInk, line width=0.6pt] (7pt,0) -- (14pt,0); \node[lvstart] at (0,0) {}; \node[lvend] at (15pt,0) {};}}%
\sbox{\lvGlyphBag}{\tikz[baseline=-0.6ex]{\node[lvbag] at (0,0) {};}}%
\sbox{\lvGlyphTarget}{\tikz[baseline=-0.6ex]{\fill[lvTargetLo] (0,0) ellipse [x radius=6pt, y radius=3pt];%
  \fill[lvTargetHi] (0,0) ellipse [x radius=3.4pt, y radius=1.7pt];}}%
\sbox{\lvGlyphGood}{\tikz[baseline=-0.6ex]{\node[lvgood] at (0,0) {};}}%
\sbox{\lvGlyphBad}{\tikz[baseline=-0.6ex]{\node[lvbad] at (0,0) {};}}%
\sbox{\lvGlyphApproxOne}{\tikz[baseline=-0.6ex]{\draw[lvTeal, line width=0.9pt] (0,0)%
  ellipse [x radius=6pt, y radius=3pt];}}%
\sbox{\lvGlyphDrift}{\tikz[baseline=-0.6ex]{\draw[lvInk, line width=0.8pt]%
  (0,0) -- (15pt,0); \node[lvstart] at (0,0) {};}}%
\sbox{\lvGlyphEnds}{\tikz[baseline=-0.6ex]{\foreach \p in {(0,0.6pt),(1.8pt,-1pt),(3.4pt,1.1pt),(4.6pt,-0.4pt),(6.2pt,0.7pt)}%
  \fill[lvGray, opacity=0.5] \p circle (0.55pt);}}%
\sbox{\lvGlyphDraws}{\tikz[baseline=-0.6ex]{\foreach \p in {(0,0.5pt),(2.2pt,-1.2pt),(4pt,1pt),(5.5pt,-0.6pt)}%
  \fill[lvOrange, opacity=0.7] \p circle (0.6pt);}}%
\begin{tikzpicture}
\pgfplotsset{
  colormap={lvpath}{color=(lvPathLo) color=(lvInk)},
  lvpanel/.style={scale only axis, width=1.2600in, height=1.2600in, anchor=south west, axis on top, clip=true,
    enlargelimits=false, xmin=-2.1000, xmax=0.9000, ymin=-0.9000, ymax=2.1000, xtick=\empty, ytick=\empty,
    axis line style={line width=0.5pt, lvInk}}}

\begin{axis}[lvpanel, at={(0.0000in,0in)}]
  \addplot graphics[xmin=-2.1000, xmax=0.9000, ymin=-0.9000, ymax=2.1000] {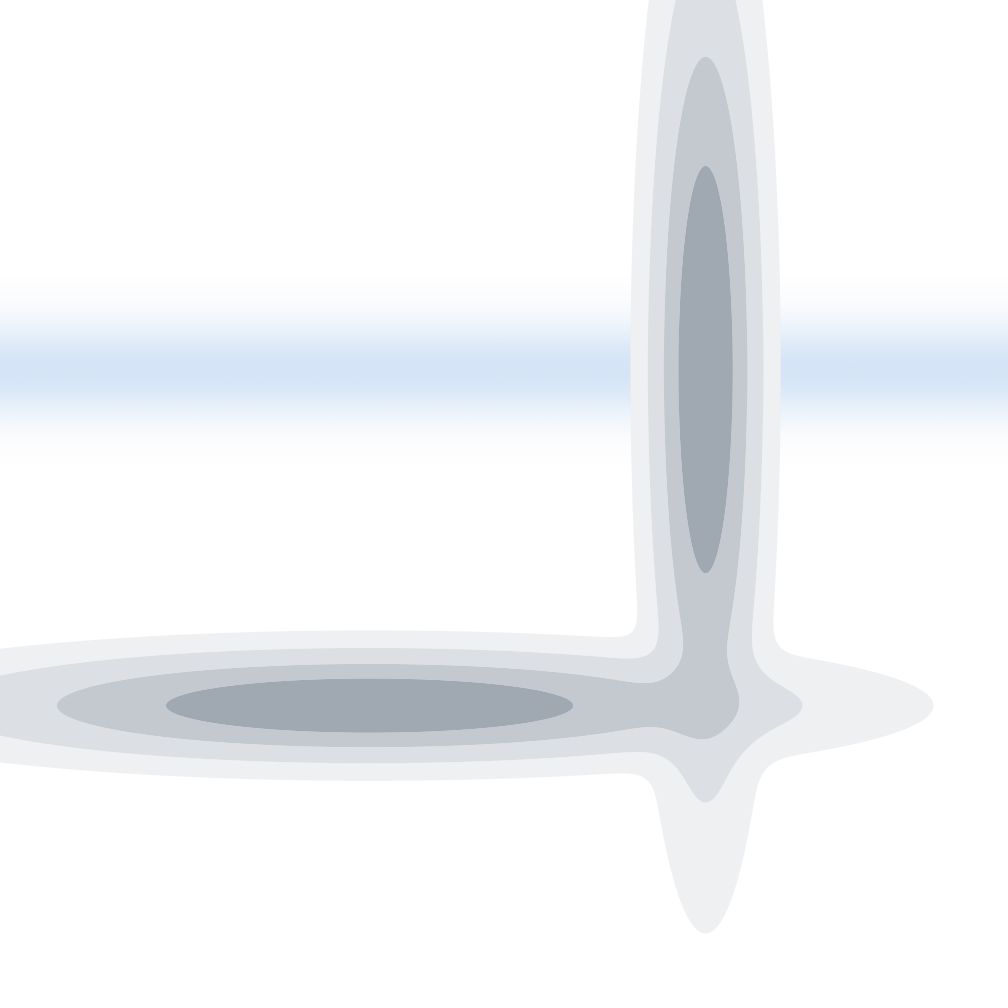};
  \draw[lvLik, line width=0.5pt, opacity=0.8] (axis cs:-2.1000,1.0000) -- (axis cs:0.9000,1.0000);
  \draw[lvTeal, line width=0.9pt, opacity=0.90] (axis cs:-0.4971,0.0114) ellipse [x radius=0.4000, y radius=0.4000];
  \addplot[lvInk, line width=0.8pt, line join=round]
    table[x=x, y=y, col sep=comma] {figures/gmm-toy-exp/data/path-smooth-daps.csv};
  \node[lvstart] at (axis cs:-0.4971,0.0114) {};
  \node[lvbad] at (axis cs:-0.8880,1.0737) {};
\end{axis}

\begin{axis}[lvpanel, at={(1.3400in,0in)}]
  \addplot graphics[xmin=-2.1000, xmax=0.9000, ymin=-0.9000, ymax=2.1000] {figures/gmm-toy-exp/img/background.png};
  \draw[lvLik, line width=0.5pt, opacity=0.8] (axis cs:-2.1000,1.0000) -- (axis cs:0.9000,1.0000);
  \draw[lvTeal, line width=0.9pt, opacity=0.90] (axis cs:-0.4971,0.0114) ellipse [x radius=0.4000, y radius=0.4000];
  \addplot[lvInk, line width=0.8pt, line join=round]
    table[x=x, y=y, col sep=comma] {figures/gmm-toy-exp/data/path-smooth-pc-gauss.csv};
  \node[lvstart] at (axis cs:-0.4971,0.0114) {};
  \node[lvbad] at (axis cs:-0.8880,1.0737) {};
\end{axis}

\begin{axis}[lvpanel, at={(2.6800in,0in)}]
  \addplot graphics[xmin=-2.1000, xmax=0.9000, ymin=-0.9000, ymax=2.1000] {figures/gmm-toy-exp/img/background.png};
  \draw[lvLik, line width=0.5pt, opacity=0.8] (axis cs:-2.1000,1.0000) -- (axis cs:0.9000,1.0000);
  \draw[lvTeal, line width=0.9pt, opacity=0.64] (axis cs:-0.7663,0.0908) ellipse [x radius=0.3840, y radius=0.1506];
  \draw[lvTeal, line width=0.9pt, opacity=0.64] (axis cs:-0.0617,1.0360) ellipse [x radius=0.1591, y radius=0.3639];
  \addplot[lvInk, line width=0.8pt, line join=round]
    table[x=x, y=y, col sep=comma] {figures/gmm-toy-exp/data/path-smooth-bag.csv};
  \node[lvstart] at (axis cs:-0.4971,0.0114) {};
  \node[lvgood] at (axis cs:-0.1787,1.1338) {};
\end{axis}

\begin{axis}[lvpanel, at={(4.0200in,0in)}]
  \addplot graphics[xmin=-2.1000, xmax=0.9000, ymin=-0.9000, ymax=2.1000] {figures/gmm-toy-exp/img/background.png};
  \draw[lvLik, line width=0.5pt, opacity=0.8] (axis cs:-2.1000,1.0000) -- (axis cs:0.9000,1.0000);
  \draw[lvTeal, line width=0.9pt, opacity=0.79] (axis cs:-0.6192,0.0012) ellipse [x radius=0.3328, y radius=0.0784];
  \draw[lvTeal, line width=0.9pt, opacity=0.45] (axis cs:-0.0173,0.3285) ellipse [x radius=0.0784, y radius=0.3328];
  \addplot[lvInk, line width=0.8pt, line join=round]
    table[x=x, y=y, col sep=comma] {figures/gmm-toy-exp/data/path-smooth-exact.csv};
  \node[lvstart] at (axis cs:-0.4971,0.0114) {};
  \node[lvgood] at (axis cs:-0.0098,1.0750) {};
\end{axis}

\node[anchor=south, font=\footnotesize, align=center, inner sep=0pt] at (0.6300in,1.3100in) {Gaussian\\Approximation};
\node[anchor=south, font=\footnotesize, align=center, inner sep=0pt] at (1.9700in,1.3100in) {Gaussian\\Approximation};
\node[anchor=south, font=\footnotesize, align=center, inner sep=0pt] at (3.3100in,1.3100in) {Gaussian Mixture\\Approximation};
\node[anchor=south, font=\footnotesize, align=center, inner sep=0pt] at (4.6500in,1.3100in) {$p(x_0\,|\,x_t)$};
\node[anchor=south, font=\footnotesize, inner sep=0pt] at (0.6300in,1.6550in) {DAPS};
\draw[lvGray, line width=0.4pt] (0.0600in,1.6250in) -- (1.2000in,1.6250in);
\node[anchor=south, font=\footnotesize, inner sep=0pt] at (2.6400in,1.6550in) {PC-DAPS};
\draw[lvGray, line width=0.4pt] (1.4000in,1.6250in) -- (3.8800in,1.6250in);
\node[anchor=south, font=\footnotesize, inner sep=0pt] at (4.6500in,1.6550in) {Exact};
\draw[lvGray, line width=0.4pt] (4.0800in,1.6250in) -- (5.2200in,1.6250in);
\node[font=\footnotesize] at (2.6400in,-0.1in) {$x_1$};
\node[font=\footnotesize, rotate=90] at (-0.1in,0.6300in) {$x_2$};
\node[font=\scriptsize, text=lvLik, anchor=south west, inner sep=1pt] at (0.0200in,0.8080in) {$x_2 = y$};

\node[anchor=north, font=\scriptsize, inner sep=0pt, align=center] at (2.6400in,-0.2in) {%
  \usebox{\lvGlyphPrior}~prior $p(x_0)$\hspace{9pt}%
  \usebox{\lvGlyphLik}~likelihood $p(y\,|\,x_0)$\hspace{9pt}%
  \usebox{\lvGlyphApproxOne}~approximation of $p(x_0\,|\,x_t)$\\[3pt]%
  \usebox{\lvGlyphDrift}~Langevin run from $\hat{x}_0$\hspace{9pt}%
  \usebox{\lvGlyphGood}~ends on the prior\hspace{9pt}%
  \usebox{\lvGlyphBad}~ends off the prior};
\end{tikzpicture}

%% file: tables/implementation-ablation.tex
\begin{table*}[t]
\caption{\textbf{Comparison of the ODE schedule and final Langevin refinement.} Ablated over 10 randomly selected FFHQ images using the PC-DAPS-1K configuration in phase retrieval task.}
\label{tab:implementation-ablation}
\centering
\small
\setlength{\tabcolsep}{7pt}
\renewcommand{\arraystretch}{1.15}
\begin{tabular}{llcccccc}
\toprule
\multirow{2}{*}{\textbf{ODE steps}} &
\multirow{2}{*}{\textbf{Final Langevin}} &
\multicolumn{3}{c}{\textbf{Mean reconstruction}} &
\multicolumn{3}{c}{\textbf{Oracle selection}} \\
\cmidrule(lr){3-5}\cmidrule(lr){6-8}
& & \textbf{PSNR $\uparrow$} & \textbf{SSIM $\uparrow$} &
\textbf{LPIPS $\downarrow$} & \textbf{PSNR $\uparrow$} &
\textbf{SSIM $\uparrow$} & \textbf{LPIPS $\downarrow$} \\
\midrule
\multirow{2}{*}{5} & Included
& 29.50 & 0.787 & 0.168 & 29.63 & 0.789 & 0.167 \\
& Dropped
& 30.01 & 0.831 & 0.144
& 30.14 & 0.833 & 0.143 \\
\addlinespace
\multirow{2}{*}{1} & Included
& 25.02 & 0.692 & 0.274 & 28.43 & 0.780 & 0.188 \\
& Dropped
& 25.41 & 0.739 & 0.254 & 28.93 & 0.831 & 0.164 \\
\addlinespace
\multirow{2}{*}{$1{\to}5$} & Included
& 28.11 & 0.774 & 0.181 & 28.34 & 0.778 & 0.179 \\
& Dropped
& 29.83 & 0.818 & 0.158 & 29.99 & 0.822 & 0.149 \\
\bottomrule
\end{tabular}
\end{table*}

%% file: tables/phase-retrieval-mean-of-k.tex
\begin{table*}[t]
\centering
\caption{\textbf{Phase retrieval under a mean-of-$k$ evaluation.} All methods are evaluated using the mean performance across $k$ reconstructions, rather than selecting the reconstruction closest to the ground truth. This exposes failures that best-of-$k$ evaluation can mask, since individual reconstructions may be measurement-consistent but differ from the ground truth due to color inversion, rotation, or translation. PC-DAPS achieves higher reconstruction quality compared to other baselines.}
\label{tab:phase-retrieval-mean-of-k}
\resizebox{\textwidth}{!}{%
\begin{tabular}{l!{\vrule}l!{\vrule}ccc!{\vrule}ccc}
\toprule
\multirow{2}{*}{\textbf{Task}} & \multirow{2}{*}{\textbf{Method}} & \multicolumn{3}{c}{\textbf{FFHQ}} & \multicolumn{3}{c}{\textbf{ImageNet}} \\  

 &  &  \textbf{PSNR ($\uparrow$)} & \textbf{SSIM ($\uparrow$)} & \textbf{LPIPS ($\downarrow$)} & \textbf{PSNR ($\uparrow$)} & \textbf{SSIM ($\uparrow$)} & \textbf{LPIPS ($\downarrow$)} \\ \midrule

\multirow{6}{*}{Phase retrieval} 
 & PC-DAPS & $\textbf{30.09}_{\pm 1.77}$ & $\textbf{0.834}_{\pm 0.089}$ & $\textbf{0.152}_{\pm 0.088}$ & $\textbf{21.75}_{\pm 6.10}$ & $\textbf{0.469}_{\pm 0.132}$ & $\textbf{0.373}_{\pm 0.128}$ \\

 & DAPS & $25.11_{\pm 5.73}$ & $0.666_{\pm 0.141}$ & $0.283_{\pm 0.138}$ & $15.70_{\pm 3.90}$ & $0.317_{\pm 0.153}$ & $0.516_{\pm 0.101}$ \\
 & DPS & $14.13_{\pm 2.72}$ & $0.398_{\pm 0.106}$ & $0.508_{\pm 0.085}$ & $13.38_{\pm 4.03}$ & $0.311_{\pm 0.208}$ & $0.560_{\pm 0.123}$ \\
 & PnP-DM  & $17.68_{\pm 5.50}$ & $0.458_{\pm 0.191}$ & $0.523_{\pm 0.149}$ & $14.80_{\pm 5.60}$ & $0.382_{\pm 0.170}$ & $0.570_{\pm 0.110}$ \\
 & DPnP & $16.57_{\pm 6.76}$ & $0.479_{\pm 0.213}$ & $0.499_{\pm 0.193}$ & $14.69_{\pm 3.61}$ & $0.427_{\pm 0.180}$ & $0.584_{\pm 0.101}$ \\
 & RED-Diff & $14.30_{\pm 3.90}$ & $0.347_{\pm 0.180}$ & $0.630_{\pm 0.095}$
& $13.90_{\pm 3.30}$ & $0.354_{\pm 0.070}$ & $0.558_{\pm 0.105}$ \\
\bottomrule

\end{tabular}
}
\end{table*}

%% file: tables/sampling-configurations.tex
 \begin{table*}[t]
  \caption{\textbf{Sampling configurations.} ODE steps, annealing steps and number of function evaluations (NFE) for the DAPS and PC-DAPS configurations
  used in the ablation studies. PC-DAPS uses a bag of $K=4$ particles; $1{\to}k$ means one Euler step for the first $65\%$ of the annealing steps and $k$
  steps afterwards. NFE is the number of denoiser evaluations per sample (per run for DAPS, per particle for PC-DAPS), counted during sampling. PC-DAPS
  configurations are named after the DAPS configuration with the same annealing steps and final ODE steps; they use fewer ODE steps early on but one extra
  denoiser evaluation per annealing step to fit the mixture, so their NFE differs from the name.}
  \label{tab:sampling-configurations}
  \centering
  \small
  \setlength{\tabcolsep}{7pt}
  \renewcommand{\arraystretch}{1.12}
  \begin{adjustbox}{max width=\textwidth}
  \begin{tabular}{@{}l!{\vrule width 0.5pt}ccccccc@{}}
  \toprule
  \textbf{Configuration} & \textbf{DAPS-50} & \textbf{DAPS-100} & \textbf{DAPS-200} & \textbf{DAPS-400} & \textbf{DAPS-1K} & \textbf{DAPS-2K} &
  \textbf{DAPS-4K} \\
  \midrule
  \textbf{ODE Steps}       & 2  & 2  & 2   & 4   & 5   & 8    & 10   \\
  \textbf{Annealing Steps} & 25 & 50 & 100 & 100 & 200 & 250  & 400  \\
  \textbf{NFE}             & 50 & 100 & 200 & 400 & 1000 & 2000 & 4000 \\
  \midrule
  \textbf{Configuration} & \textbf{PC-DAPS-50} & \textbf{PC-DAPS-100} & \textbf{PC-DAPS-200} & \textbf{PC-DAPS-400} & \textbf{PC-DAPS-1K} & & \\
  \midrule
  \textbf{ODE Steps}       & $1{\to}2$ & $1{\to}2$ & $1{\to}2$ & $1{\to}4$ & $1{\to}5$ & & \\
  \textbf{Annealing Steps} & 25 & 50  & 100 & 100 & 200 & & \\
  \textbf{NFE}             & 66 & 131 & 261 & 331 & 732 & & \\
  \bottomrule
  \end{tabular}
  \end{adjustbox}
  \end{table*}